\documentclass[a4paper,fleqn]{cas-dc}

\usepackage{times}  
\usepackage{helvet}  
\usepackage{courier}  
\usepackage{graphicx} 
\usepackage{natbib}  
\usepackage{caption} 

\usepackage{algorithm}
\usepackage{algorithmic}
\usepackage{float}
\usepackage{booktabs}
\usepackage{amsmath}
\usepackage{amssymb}
\usepackage{multirow} 
\usepackage{pifont}
\usepackage[table]{xcolor}
\usepackage{makecell} 
\usepackage{newfloat}
\usepackage{listings}
\usepackage{rotating}
\usepackage{graphicx}
\usepackage{subcaption}

\begin{document}
\let\WriteBookmarks\relax
\def\floatpagepagefraction{1}
\def\textpagefraction{.001}

\shorttitle{Rethinking Patch Based Multivariate Time Series Forecasting with Semantic Structured Partitioning}    

%

\author[1]{Zhiquan Huang}
\fnmark[1]  
\ead{zqhuang@stu.haust.edu.cn}

\author[2]{Jiazhe Wang}
\fnmark[1] 
\ead{231451080320@stu.haust.edu.cn}

\affiliation[1]{organization={School of Information Engineering},
            addressline={Henan University of Science and Technology},
            city={Luoyang},
            postcode={471023},
            state={Henan},
            country={China}}

\affiliation[2]{organization={School of Software},
            addressline={Henan University of Science and Technology},
            city={Luoyang},
            postcode={471000},
            state={Henan},
            country={China}}

\author[2]{Linjing Xue} 
\ead{ljxue@stu.haust.edu.cn}  

\author[1]{Ming Liu} 
\ead{liuming@haust.edu.cn}  

\author[1]{Meiwen Li} 
\ead{mwli@stu.haust.edu.cn}

\author[1]{Ruijuan Zheng\corref{cor1}} 
\cormark[1]  
\ead{zhengruijuan@haust.edu.cn}  

\cortext[cor1]{Corresponding author}
\fntext[fn1]{These authors contributed equally to this work.}

\title [mode = title]{Rethinking Patch Based Multivariate Time Series Forecasting with Semantic Structured Partitioning}  

\begin{abstract}
Multivariate time series forecasting (MTSF) is a fundamental task in many real world applications. Existing patch based methods typically rely on fixed, multiscale, or extendable partitioning schemes. However, these strategies primarily improve partitioning flexibility without explicitly modeling the semantic coherence of local temporal patterns or the higher order dependencies among them. To address this limitation, we propose SCPaT, a unified framework for modeling temporal sequences from a semantic structure perspective.
Specifically, SCPaT adaptively decomposes input sequences into semantically coherent units and constructs a dynamic semantic graph to model directed dependencies among them, from which higher order semantic blocks are identified. Based on these structured representations, an importance aware routing mechanism adaptively assigns semantic blocks to a subset of experts for tailored modeling. Extensive experiments on 12 real world datasets demonstrate that SCPaT achieves strong performance in both long and short horizon forecasting.

\end{abstract}






\begin{keywords}
Multivariate time series forecasting \sep Semantic structured partitioning\sep Transfer entropy graph\sep Importance aware routing
\end{keywords}

\maketitle

\section{Introduction}

Multivariate time series forecasting (MTSF) plays an important role in various domains~\citep{qiu2024tfb,zhou2025dual}, including electricity load forecasting~\citep{gasparin2022deep}, weather forecasting~\citep{karevan2020transductive}, and traffic flow forecasting~\citep{lippi2013short}. 
Recent advances in deep learning have led to a wide range of forecasting architectures for MTSF~\citep{huang2025endexformer}.
Among them, patch based modeling has emerged as an effective paradigm for capturing complex temporal dependencies~\citep{cao2025mspatch}. 
By partitioning long sequences into temporal patches, these methods reduce modeling complexity while preserving fine grained local patterns~\citep{zhang2023crossformer}. 
However, the effectiveness of these models hinges critically on how the input sequence is partitioned, a design choice that, as we will show, has profound implications for semantic preservation. Consequently, the design of patch partitioning strategies has become a key consideration in patch based Multivariate time series forecasting .

\begin{figure}
\centering
\includegraphics[width=0.65\columnwidth]{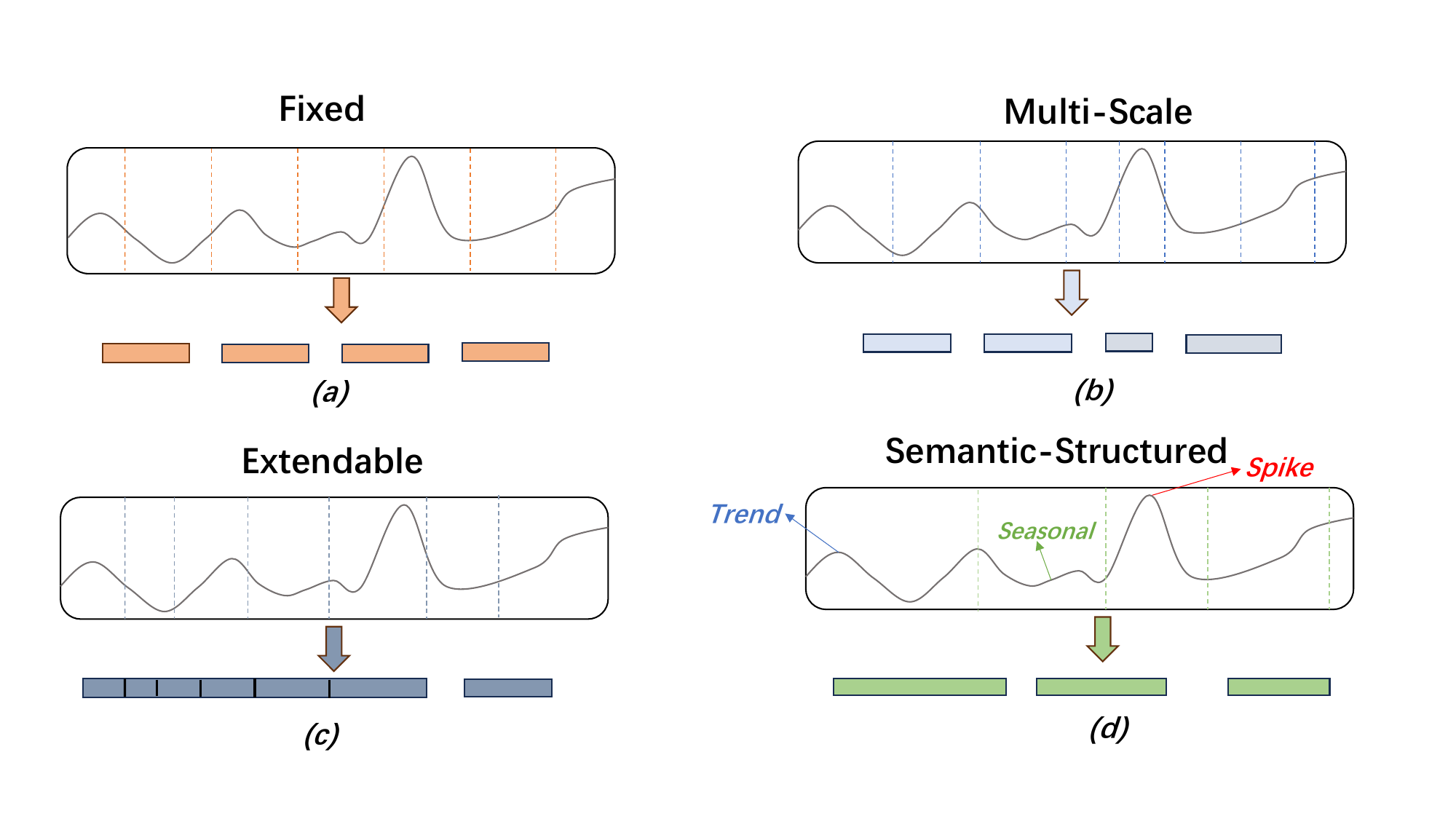}
\caption{Comparison of four patch based schemes: fixed partitioning, multi-scale partitioning, extendable partitioning, and the proposed semantic structured partitioning.
}
\label{fig1}
\end{figure}

Existing patch partitioning strategies can be broadly categorized into three types: fixed partitioning~\citep{patchtst}, multi-scale partitioning~\citep{wu2022timesnet}, and extendable partitioning~\citep{huang2024hdmixer}. 
Among these strategies, fixed partitioning is the most widely adopted in patch based forecasting. 
It divides a time series into equal length segments and treats each segment as an input unit, thereby reducing the effective sequence length and simplifying the modeling process. 
As illustrated in Figure~\ref{fig1}(a), this design is computationally efficient and performs well on time series with regular and predictable patterns. 
However, because it relies on uniform segmentation, fixed partitioning often fails to align with meaningful temporal boundaries in nonstationary sequences, particularly in the presence of trend shifts or abrupt events~\citep{kazemi2020assessing}. 
This limitation restricts its ability to represent complex temporal dynamics and motivates the development of more flexible partitioning strategies.

To overcome the limited flexibility of fixed partitioning, researchers have proposed multi-scale partitioning. 
This strategy partitions the time series at multiple levels of granularity, so that scale specific representations can be modeled separately and then integrated.
As shown in Figure~\ref{fig1}(b), this strategy provides a richer characterization of temporal dynamics by capturing both local short term fluctuations and global long term trends~\citep{ye2023multi}. 
However, when similar temporal patterns are captured at multiple granularities, the corresponding representations may overlap substantially across scales, thereby introducing redundancy during cross scale integration~\citep{hu2025adaptive,liu2025disms}.
Such redundancy may in turn encourage the model to learn superfluous features. 
This problem becomes more pronounced in datasets with complex and overlapping dependencies, where heterogeneous dynamic patterns are more difficult to disentangle.
These limitations motivate the development of more adaptive partitioning mechanisms.

Extendable partitioning further increases flexibility by allowing the length of each temporal segment to vary according to data characteristics~\citep{zhang2025tf4tf,liu2025semantic}. 
By adapting segment granularity to local dynamics, this strategy better preserves temporal boundary information and captures local semantic patterns. 
As illustrated in Figure~\ref{fig1}(c), such adaptive partitioning provides greater flexibility for modeling the diversity and complexity of temporal variations. 
Nevertheless, existing extendable partitioning methods typically determine segmentation granularity using heuristic rules, which limits their ability to handle complex higher order dependencies. 
In scenarios characterized by strong nonlinear dependencies or multiple coexisting temporal patterns, these methods often struggle to model interactions among heterogeneous patterns, which results in suboptimal forecasting performance.
These observations suggest that improving partition flexibility alone is insufficient for modeling the semantic structure underlying complex temporal dynamics.

Despite the advances achieved by existing methods, a critical research gap remains fundamentally unaddressed: none of the current partitioning strategies explicitly accounts for the intrinsic semantic coherence of temporal segments, nor do they model the higher order structural interactions among heterogeneous temporal patterns~\cite{xu2025frequency}. In essence, all these approaches treat patch generation as a geometric preprocessing step rather than as an integral semantic modeling task. This deficiency is particularly detrimental when trends, periodicities, and abrupt fluctuations coexist in the same time series, as the model lacks a principled and feasible mechanism to disentangle these diverse components and establish structured relationships among them. Therefore, we argue that the core challenge of patch-based forecasting lies not in pursuing greater partition flexibility, but in constructing a representation system that embeds semantic structure.

Based on the above analysis, we propose SCPaT, a Transformer based framework for MTSF that models dynamic semantic structure through semantic structured partitioning. 
Specifically, SCPaT extracts fundamental temporal components from raw time series and organizes them into hierarchical semantic blocks, which provide multilevel semantic abstractions for forecasting. 
These semantic blocks are then incorporated into the attention mechanism through an importance aware routing module, which enables the model to emphasize informative representations and capture dependencies across different semantic types. 
Extensive experiments on multiple benchmarks show that SCPaT achieves state of the art results in both long term and short term forecasting tasks. The main contributions of this work are summarized as follows:

\begin{itemize}
    \item \textbf{Systematic Analysis.} We present a systematic analysis of three representative patch partitioning strategies for time series forecasting and show that their limitations reflect the lack of an explicit semantic structured partitioning mechanism.
    
    \item \textbf{Innovative Framework.} We propose SCPaT, a Transformer based framework that integrates hierarchical semantic blocks and an importance aware routing mechanism into attention, enabling more effective modeling of key dependencies in MTSF.
    
    \item \textbf{Extensive Evaluations.} Extensive experiments on long term and short term forecasting benchmarks demonstrate that SCPaT achieves state of the art performance across a wide range of datasets.
\end{itemize}

\section{Related Work}

\subsection{Transformer based Time Series Forecasting}

Transformer-based models have become widely used in MTSF due to their strong parallel modeling capability and ability to capture long-range dependencies~\citep{qiu2026dag,ha2026tivat}. 
Representative methods have improved forecasting performance from different perspectives. 
For instance, iTransformer~\citep{itransformer} enhances cross-variable dependency modeling by reorganizing tokens along the variable dimension. 
Pathformer~\citep{chen2024pathformer} captures both local and global temporal contexts using multiscale patch partitioning with an adaptive path strategy. 
Fredformer~\citep{piao2024fredformer} mitigates high-frequency bias through frequency debiasing. 
However, these methods are still limited in their ability to explicitly characterize heterogeneous dynamics within patches and capture interactions across different temporal patterns \citep{yang2025beyond}. 
This limitation becomes particularly evident when trends, periodicity, and abrupt fluctuations coexist in the same time series~\citep{xue2026beyond}.  
Such scenarios require mechanisms that can distinguish local semantic differences and model complex dependencies more effectively.

\subsection{Patch Based Time Series Forecasting}

Patch based methods have become an important direction in time series forecasting, with early studies primarily relying on fixed partitioning strategies~\citep{wu2025catch,chen2024pathformer}. 
PatchTST~\citep{patchtst}, for example, segments time series into equal length patches to improve representation efficiency. 
However, fixed partitioning often fails to preserve meaningful temporal boundaries, which may lead to the loss of salient local patterns and semantic inconsistency within patches. 
To address this issue, subsequent studies have explored multi-scale partitioning strategies. 
For instance, TimesNet~\citep{wu2022timesnet} models temporal patterns across different scales, while FEDformer~\citep{zhou2022fedformer} integrates local and global trends under a frequency domain decomposition framework. 
Extendable partitioning has further emerged as a flexible alternative, with HDMixer~\citep{huang2024hdmixer} introducing length adaptive patches that dynamically adjust segment boundaries, improving semantic integrity and modeling flexibility. 
Despite these advances, existing patch based methods remain limited in their ability to explicitly capture fine grained semantic variation and higher order temporal dependencies. All of them formulate patch generation as a geometric partitioning problem, whereas our SCPaT redefines it from a semantic structure perspective.

\begin{figure*}
\centering
\includegraphics[width=0.98\textwidth]{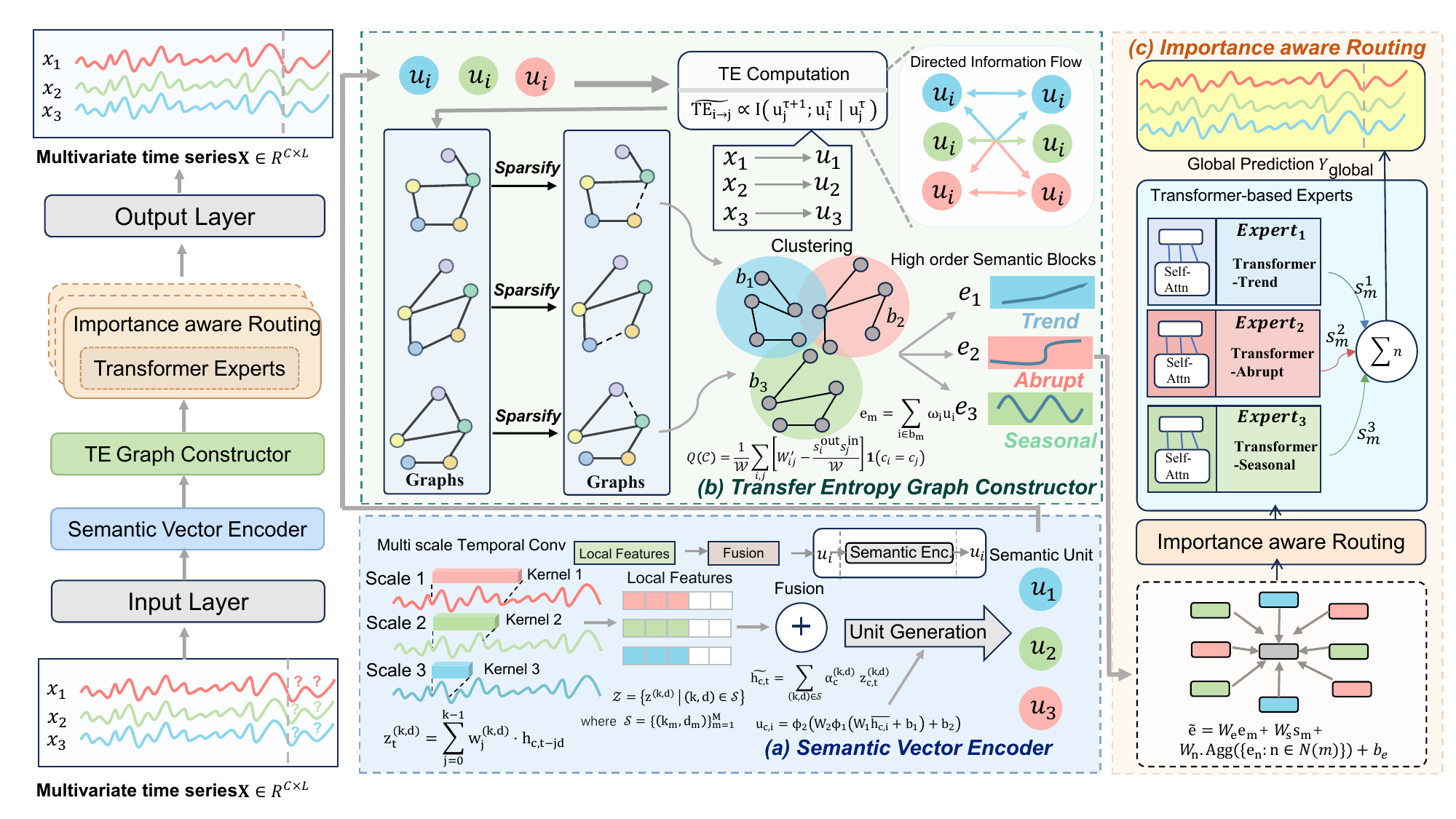} 
\caption{SCPaT architecture consists of: \textit{(a)}~ Semantic Vector Encoder, which encodes time series into semantic units; \textit{(b)}~Transfer Entropy Graph Constructor, quantifies directed dependencies and constructs a dynamic semantic graph; and \textit{(c)}~Importance aware Routing, allocates customized processing strategies to different semantic blocks for effective modeling.}
\label{fig2}
\end{figure*}


\section{Proposed Methodology}

\subsection{Preliminaries}

Given a multivariate time series with $C$ variables and a historical look-back window of length $L$, we denote the input as $\mathbf{X}=[\mathbf{x}_1^\top,\mathbf{x}_2^\top,\dots,\mathbf{x}_C^\top]^\top\in\mathbb{R}^{C\times L}$, where $\mathbf{x}_c\in\mathbb{R}^{L}$ represents the historical observations of the $c$-th variable. The objective of multivariate time series forecasting is to learn a parameterized mapping $f_{\Theta}(\cdot)$ that predicts future observations over a forecasting horizon of length $H$, formulated as $\hat{\mathbf{Y}}=f_{\Theta}(\mathbf{X})$, where $\hat{\mathbf{Y}}\in\mathbb{R}^{C\times H}$. Instead of partitioning each variable into predefined fixed-length patches, SCPaT adaptively partitions the historical sequence into semantic units according to local temporal characteristics. These semantic units are used to construct a directed dependency graph, from which higher-order semantic blocks are identified. The resulting blocks are then processed by the importance-aware routing module and specialized experts to generate the final forecasts.

\subsection{Semantic Vector Encoder}

The Semantic Vector Encoder transforms the input sequence 
\(\mathbf{X}\in\mathbb{R}^{C\times L}\) into a collection of local semantic representations. 
It first extracts temporal patterns under multiple receptive fields, then integrates the resulting scale-specific features, and finally partitions the fused representations into semantic units according to local temporal characteristics.

\subsubsection{Multi-scale Temporal Convolution}

To characterize local temporal dynamics at different resolutions, we employ multiple temporal convolution branches over the projected feature sequence. For the \(c\)-th variable, let \(\mathbf{h}_c=\{\mathbf{h}_{c,t}\}_{t=1}^{L}\) denote its projected feature sequence, where \(\mathbf{h}_{c,t}\in\mathbb{R}^{d_h}\) denotes the \(d_h\)-dimensional latent representation at time step \(t\). For a branch with kernel size \(k\) and dilation factor \(d\), the output at time step \(t\) is computed as follows:
\begin{equation}
\mathbf{z}_{c,t}^{(k,d)}
=
\sum_{j=0}^{k-1}
w_{j}^{(k,d)}\,\mathbf{h}_{c,t-jd},
\label{eq:ms_conv}
\end{equation}
where \(\mathbf{z}_{c,t}^{(k,d)}\in\mathbb{R}^{d_h}\) is the representation extracted under configuration \((k,d)\), and \(w_{j}^{(k,d)}\) denotes the \(j\)-th kernel coefficient. Zero padding is applied when \(t-jd<1\) to preserve the sequence length.

The branch output can be written compactly as follows:
\begin{equation}
\mathbf{Z}_{c}^{(k,d)}
=
\mathbf{W}^{(k,d)} * \mathbf{H}_{c},
\label{eq:ms_conv_compact}
\end{equation}
where \(*\) denotes temporal convolution, 
\(\mathbf{H}_{c}=[\mathbf{h}_{c,1},\dots,\mathbf{h}_{c,L}]\) represents the complete projected feature sequence of the \(c\)-th variable, and 
\(\mathbf{W}^{(k,d)}\) denotes the convolution kernel associated with the temporal configuration \((k,d)\). 
The corresponding effective receptive field is \(r^{(k,d)}=1+(k-1)d\), jointly determined by the kernel size and dilation factor.

We consider \(M\) convolution branches with different temporal configurations. 
These configurations are collected in 
\(\mathcal{S}=\{(k_m,d_m)\}_{m=1}^{M}\), 
where each pair \((k_m,d_m)\) specifies the kernel size and dilation factor of the \(m\)-th branch. 
Applying these branches in parallel yields the following multi-scale feature set:
\begin{equation}
\mathcal{Z}_c
=
\left\{
\mathbf{Z}_{c}^{(k,d)}
\mid
(k,d)\in\mathcal{S}
\right\}.
\label{eq:ms_feature_set}
\end{equation}
These representations are subsequently fused to obtain a unified multi-scale representation.

\subsubsection{Adaptive Multi-scale Fusion}

The temporal branches capture complementary patterns under different receptive fields. To integrate these scale-specific representations, we perform learnable multi-scale fusion over the branch set \(\mathcal{S}\). The fused representation of variable \(c\) at time step \(t\) is defined as:
\begin{equation}
\tilde{\mathbf{h}}_{c,t}
=
\sum_{(k,d)\in\mathcal{S}}
\alpha_{c}^{(k,d)}\,\mathbf{z}_{c,t}^{(k,d)},
\label{eq:fusion_main}
\end{equation}
where \(\tilde{\mathbf{h}}_{c,t}\in\mathbb{R}^{d_h}\) denotes the fused representation at time step \(t\), and \(\alpha_{c}^{(k,d)}\) denotes the fusion weight of configuration \((k,d)\) for variable \(c\).

The fusion weights are obtained by normalizing learnable branch scores as follows:
\begin{equation}
\alpha_{c}^{(k,d)}
=
\frac{\exp\!\big(\gamma_{c}^{(k,d)}\big)}
{\sum_{(k',d')\in\mathcal{S}}\exp\!\big(\gamma_{c}^{(k',d')}\big)},
\label{eq:fusion_weight}
\end{equation}
where \(\gamma_{c}^{(k,d)}\) denotes the trainable score associated with configuration \((k,d)\) for variable \(c\). This variable-specific weighting allows the encoder to learn distinct scale preferences for different variables while aggregating complementary temporal information. The resulting fused sequence is denoted by \(\tilde{\mathbf{H}}_c=[\tilde{\mathbf{h}}_{c,1},\tilde{\mathbf{h}}_{c,2},\dots,\tilde{\mathbf{h}}_{c,L}]\).

\subsubsection{Semantic Unit Generation}

Given the fused sequence \(\tilde{\mathbf{H}}_c\), we partition each variable into a sequence of semantic units with adaptive lengths and encode each unit into a compact representation. A unit starting at position \(i\) with length \(l_i\) covers the interval \([i,i+l_i-1]\), and the subsequent unit begins at \(i+l_i\).
To adapt the partition to local temporal dynamics, the unit length is selected from the candidate set \(\mathcal{L}=\{1,\dots,L_{\max}\}\), where \(L_{\max}\) denotes the maximum unit length. For a candidate segment \(\tilde{\mathbf{H}}_c[i:i+l'-1]\), we first summarize the latent representation at each time step by \(a_{c,t}=\frac{1}{d_h}\sum_{r=1}^{d_h}\tilde{h}_{c,t,r}\), where \(\tilde{h}_{c,t,r}\) denotes the \(r\)-th dimension of \(\tilde{\mathbf{h}}_{c,t}\). The local variation is then quantified as follows:
\begin{equation}
\mathrm{Var}\!\left(\tilde{\mathbf{H}}_c[i:i+l'-1]\right)
=
\frac{1}{l'}
\sum_{t=i}^{i+l'-1}
\left(
a_{c,t}-\bar{a}_{c,i,l'}
\right)^2,
\label{eq:local_var_final}
\end{equation}
where \(\bar{a}_{c,i,l'}=\frac{1}{l'}\sum_{t=i}^{i+l'-1}a_{c,t}\) denotes the mean activation of the candidate segment.

We further define the admissible length set as \(\mathcal{L}_i^{\delta}=\{l'\in\mathcal{L}\mid \mathrm{Var}(\tilde{\mathbf{H}}_c[i:i+l'-1])>\delta\}\), where \(\delta\) controls the sensitivity of boundary detection to local temporal variation. The unit length is determined as follows:
\begin{equation}
l_i
=
\begin{cases}
\min \mathcal{L}_i^{\delta}, & \text{if } \mathcal{L}_i^{\delta}\neq\emptyset,\\[4pt]
L_{\max}, & \text{otherwise}.
\end{cases}
\label{eq:unit_length_final}
\end{equation}
This criterion assigns longer units to stable regions and shorter units to rapidly varying regions, while the semantic content is preserved by the learned temporal representations.
The resulting units may have different lengths. We therefore summarize each segment into a fixed dimensional representation through mean pooling, given by \(\bar{\mathbf{h}}_{c,i}=\frac{1}{l_i}\sum_{t=i}^{i+l_i-1}\tilde{\mathbf{h}}_{c,t}\). The corresponding semantic unit representation is then obtained as follows:
\begin{equation}
\mathbf{u}_{c,i}
=
\phi_2\!\left(
\mathbf{W}_2
\phi_1\!\left(
\mathbf{W}_1\bar{\mathbf{h}}_{c,i}+\mathbf{b}_1
\right)
+\mathbf{b}_2
\right),
\label{eq:semantic_unit_final}
\end{equation}
where \(\mathbf{W}_1\) and \(\mathbf{W}_2\) are learnable projection matrices, \(\mathbf{b}_1\) and \(\mathbf{b}_2\) are learnable bias vectors, and \(\phi_1(\cdot)\) and \(\phi_2(\cdot)\) denote nonlinear activation functions.

Independent semantic unit generation across variables may result in different numbers of units and temporal spans. We therefore align units according to their temporal coverage and denote the resulting sequence for variable \(c\) as \(\mathcal{U}_c=\{\mathbf{u}_{c,1},\mathbf{u}_{c,2},\dots,\mathbf{u}_{c,N_c}\}\). All units are further reindexed into a unified set \(\mathcal{U}=\{\mathbf{u}_n\}_{n=1}^{N_u}\), where \(N_u=\sum_{c=1}^{C}N_c\), for subsequent dependency modeling.

\subsection{Transfer Entropy Graph Constructor}

The transfer entropy graph constructor estimates directed dependencies among semantic units and organizes them into a dynamic semantic graph for higher order semantic block construction. By incorporating directional and conditional dependencies, the resulting graph characterizes nonlinear and asymmetric interactions beyond conventional correlation based relations.

\subsubsection{Transfer Entropy Computation}

Conventional correlation based measures are typically symmetric and limited in representing directed conditional dependencies. We therefore introduce a differentiable surrogate inspired by transfer entropy to model directed dependence between aligned semantic units.
For an aligned source and target pair \((i,j)\), let \(\mathbf{u}_i^{\tau}\) and \(\mathbf{u}_j^{\tau}\) denote their representations at the aligned transition step \(\tau\), where \(i\) and \(j\) identify the source and target units, respectively. We define the following conditional dependence target:
\begin{equation}
\widetilde{TE}_{i\to j}
\propto
I\!\left(
\mathbf{u}_j^{\tau+1};\mathbf{u}_i^{\tau}\mid \mathbf{u}_j^{\tau}
\right),
\label{eq:te_surrogate_target}
\end{equation}
where \(I(\cdot;\cdot\mid\cdot)\) denotes conditional mutual information, and \(\tau\) indexes valid aligned unit transitions. This quantity characterizes the additional predictive information provided by \(\mathbf{u}_i^{\tau}\) beyond the target history \(\mathbf{u}_j^{\tau}\).

To obtain a practical differentiable score, we use \(b\) to index samples in the mini batch and \(\tau\) to index aligned transitions within each sample. For each ordered pair \((i,j)\), let \(\Omega_{ij}\) denote the set of valid transitions \((b,\tau)\) for which \(\mathbf{u}_{i}^{(b,\tau)}\), \(\mathbf{u}_{j}^{(b,\tau)}\), and \(\mathbf{u}_{j}^{(b,\tau+1)}\) are all available, and let \(M_{ij}=|\Omega_{ij}|\). For \(M_{ij}>0\), the directed dependency score is defined as follows:
\begin{equation}
\widehat{TE}_{i\to j}
=
\frac{1}{M_{ij}}
\sum_{(b,\tau)\in\Omega_{ij}}
\psi\!\left(
\mathbf{u}_{j}^{(b,\tau+1)},
\mathbf{u}_{i}^{(b,\tau)},
\mathbf{u}_{j}^{(b,\tau)}
\right),
\label{eq:te_surrogate_final}
\end{equation}
where \(\psi(\cdot)\) is a learnable transition level scoring function shared across all ordered pairs.

For each valid transition \((b,\tau)\in\Omega_{ij}\), the three aligned representations are concatenated into the triplet representation \(\mathbf{q}_{ij}^{(b,\tau)}\). The scoring function \(\psi(\cdot)\) is implemented by \(\mathrm{MLP}_{\psi}\), whose input, hidden, and output dimensions are \(3d_h\), \(d_h\), and \(1\), respectively. The network maps each triplet to a scalar dependency score with parameters shared across all ordered pairs. For brevity, \(\psi(\mathbf{q}_{ij}^{(b,\tau)})\) below denotes the score produced by \(\mathrm{MLP}_{\psi}\) from the corresponding triplet representation.

To characterize the consistency of dependency scores across valid transitions, we compute their sample standard deviation. For \(M_{ij}>1\), the score dispersion is defined as follows:
\begin{equation}
\widehat{\sigma}_{i\to j}
=
\sqrt{
\frac{1}{M_{ij}-1}
\sum_{(b,\tau)\in\Omega_{ij}}
\left(
\psi\!\left(\mathbf{q}_{ij}^{(b,\tau)}\right)
-
\widehat{TE}_{i\to j}
\right)^2
},
\label{eq:te_std_final}
\end{equation}
where a larger \(\widehat{\sigma}_{i\to j}\) indicates greater variation in the directed dependency scores.
The scoring network \(\mathrm{MLP}_{\psi}\) is jointly trained with the forecasting backbone and produces relative dependency scores for graph construction.

\subsubsection{Graph Construction with Sparsification}

Based on the estimated directed dependency scores, we construct a dynamic semantic graph \(G=(V,E,W)\), where \(V\) denotes the set of semantic units, \(E\) is the directed edge set, and \(W\in\mathbb{R}^{|V|\times |V|}\) is the weighted adjacency matrix. The directed edge weight \(W_{ij}\) is defined as:
\begin{equation}
W_{ij}
=
\max\!\left(
0,\,
\widehat{TE}_{i\to j}-\eta\,\widehat{\sigma}_{i\to j}
\right),
\label{eq:edge_weight_final}
\end{equation}
where \(\eta\) controls the penalty on unstable dependency scores.

As the number of semantic units increases, the resulting adjacency matrix may become dense. We therefore sparsify the graph to suppress marginal interactions and reduce the cost of subsequent graph operations. Sparsification is performed after computing all pairwise dependency scores and thus does not reduce the pairwise computation complexity. Specifically, for each node, we retain the \(K_g\) outgoing edges with the largest weights, where \(K_g=\lceil\alpha\cdot N\rceil\) and \(N=|V|\) denotes the number of semantic units:
\begin{equation}
W'_{ij}
=
\begin{cases}
W_{ij}, & j\in \mathrm{TopK}^{\,i}_{g}(W_{i\cdot}),\\[2pt]
0, & \text{otherwise},
\end{cases}
\label{eq:sparsify_final}
\end{equation}
where \(W_{i\cdot}\) denotes the \(i\)-th row of \(W\). This operation retains the strongest outgoing dependencies while reducing graph density.
For subsequent aggregation, the sparse adjacency matrix is normalized by row as follows:
\begin{equation}
\bar{W}_{ij}
=
\frac{W'_{ij}}{\sum_{j'}W'_{ij'}+\epsilon},
\label{eq:row_norm_final}
\end{equation}
where \(\epsilon\) is a small positive constant for numerical stability.

\subsubsection{Graph Clustering for Higher order Semantic Blocks}

Given the sparse semantic graph, we perform graph clustering to identify higher order semantic blocks. Let \(c:V\rightarrow \{1,2,\dots,K_b\}\) denote the cluster assignment function, where \(K_b\) is the number of semantic blocks. The partition is obtained by maximizing directed weighted modularity, i.e., \(\mathcal{C}^{*}=\arg\max_{\mathcal{C}}Q(\mathcal{C})\), with:
\begin{equation}
Q(\mathcal{C})
=
\frac{1}{\mathcal{W}}
\sum_{i,j}
\left[
W'_{ij}
-
\frac{s_i^{\mathrm{out}}s_j^{\mathrm{in}}}{\mathcal{W}}
\right]
\mathbf{1}(c_i=c_j),
\label{eq:modularity_final}
\end{equation}
where \(s_i^{\mathrm{out}}=\sum_j W'_{ij}\), \(s_j^{\mathrm{in}}=\sum_i W'_{ij}\), \(\mathcal{W}=\sum_{i,j}W'_{ij}\), and \(\mathbf{1}(\cdot)\) denotes the indicator function. In practice, clustering is performed on the current sparse graph during each forward pass.
Each cluster defines a semantic block \(b_m=\{\mathbf{u}_i \mid c_i=m\}\), where \(m=1,2,\dots,K_b\). To obtain a compact block representation, we aggregate the semantic units within each block using attention weights. For each \(\mathbf{u}_i\in b_m\), the attention weight is defined as:
\begin{equation}
\omega_i
=
\frac{
\exp\!\big(\phi_w(\mathbf{u}_i)\big)
}{
\sum_{j\in b_m}\exp\!\big(\phi_w(\mathbf{u}_j)\big)
},
\label{eq:block_weight_final}
\end{equation}
where \(\phi_w(\cdot)\) is a lightweight scoring function. The resulting block embedding is given by \(\mathbf{e}_m=\sum_{i\in b_m}\omega_i\mathbf{u}_i\).

We further define a block statistic vector \(\mathbf{s}_m\in\mathbb{R}^{d_s}\), consisting of the normalized block size, average incoming strength, average outgoing strength, and average temporal span. These statistics provide structural cues complementary to the content representation \(\mathbf{e}_m\).
To capture block context, we construct a neighborhood graph from the sparse unit graph. Two blocks \(m\) and \(n\) are considered neighbors if at least one directed edge exists between their constituent units. Let \(\mathcal{N}(m)\) denote the neighborhood of block \(m\). The neighboring block embeddings are aggregated as follows:
\begin{equation}
\mathrm{Agg}\big(\{\mathbf{e}_n:n\in\mathcal{N}(m)\}\big)
=
\frac{1}{|\mathcal{N}(m)|}
\sum_{n\in\mathcal{N}(m)}\mathbf{e}_n,
\label{eq:neighbor_agg_final}
\end{equation}
where the aggregation is defined as the zero vector when \(\mathcal{N}(m)=\emptyset\). This aggregation provides contextual information from neighboring semantic blocks and complements the block representation for subsequent routing.

\subsection{Importance Aware Routing}
Semantic blocks in MTSF may exhibit heterogeneous temporal characteristics and require different modeling patterns. We therefore introduce an importance aware routing mechanism that adaptively assigns each block to a subset of experts based on its content, structural statistics, and neighborhood context.

\subsubsection{Expert Network}

For each semantic block \(b_m\), we first construct a routing representation by integrating its content representation, structural statistics, and neighborhood context:
\begin{equation}
\tilde{\mathbf{e}}_m
=
\mathbf{W}_e\mathbf{e}_m
+
\mathbf{W}_s\mathbf{s}_m
+
\mathbf{W}_n\,\mathrm{Agg}\big(\{\mathbf{e}_n:n\in\mathcal{N}(m)\}\big)
+
\mathbf{b}_e,
\label{eq:routing_embed_final}
\end{equation}
where \(\mathbf{W}_e\), \(\mathbf{W}_s\), and \(\mathbf{W}_n\) are learnable projection matrices, and \(\mathbf{b}_e\) is a bias vector. These projections map the three information sources into a common routing space.
Based on \(\tilde{\mathbf{e}}_m\), we further construct a semantic bias vector conditioned on the current block:
\begin{equation}
\mathbf{V}_m
=
\mathbf{V}_0
+
\beta\cdot\tanh\!\left(
\mathbf{W}_v\tilde{\mathbf{e}}_m+\mathbf{b}_v
\right),
\label{eq:semantic_bias_final}
\end{equation}
where \(\mathbf{V}_0\) is a globally learnable base vector, \(\beta\) controls the magnitude of the block dependent adjustment, and \(\mathbf{W}_v\) and \(\mathbf{b}_v\) are learnable parameters.

We employ \(R\) independently parameterized experts, denoted by
\(\{\mathrm{Expert}_k\}_{k=1}^{R}\). For subsequent expert processing, each
semantic block \(b_m\) is represented as an ordered sequence of its constituent
semantic units. We denote this sequence by
\(\mathbf{U}_m=[\mathbf{u}_{m,1},\mathbf{u}_{m,2},\dots,\mathbf{u}_{m,|b_m|}]\),
where the semantic units are arranged according to their temporal positions.
The sequence encoder \(f_k(\cdot)\) of the \(k\)-th expert maps
\(\mathbf{U}_m\) to a fixed dimensional block representation. The corresponding
expert output is defined as:
\begin{equation}
\mathbf{y}_m^{(k)}
=
\mathrm{Expert}_k(b_m;\mathbf{V}_m)
=
f_k(\mathbf{U}_m)+\mathbf{P}_k\mathbf{V}_m,
\label{eq:expert_output_final}
\end{equation}
where \(\mathbf{P}_k\) is a learnable projection matrix that maps the semantic bias vector to the representation space of expert \(k\). Thus, \(\mathbf{y}_m^{(k)}\) combines temporal features with block dependent information from its content, structure, and neighborhood.

\subsubsection{Routing Mechanism}

The router computes expert assignment scores from the routing representation of each semantic block. Specifically, the routing logits are obtained as
\(\tilde{\mathbf{s}}_m=\mathrm{MLP}_r(\tilde{\mathbf{e}}_m)\), where
\(\tilde{\mathbf{s}}_m=[\tilde{s}_m^{(1)},\dots,\tilde{s}_m^{(R)}]\)
contains the scores for all experts. The routing probability for expert \(k\) is defined as:
\begin{equation}
s_m^{(k)}
=
\frac{\exp(\tilde{s}_m^{(k)})}
{\sum_{k'=1}^{R}\exp(\tilde{s}_m^{(k')})}.
\label{eq:routing_prob_final}
\end{equation}

Rather than using a fixed Top-\(K\) rule, we adopt Top-\(P\) routing to adapt the number of active experts to the routing distribution. Specifically, experts are sorted in descending order according to \(\{s_m^{(k)}\}_{k=1}^{R}\), and the smallest set \(\mathcal{T}_m\) satisfying
\(\sum_{k\in\mathcal{T}_m}s_m^{(k)}\ge P\) is selected, where \(P\in(0,1]\).
The sparse routing weight is defined as:
\begin{equation}
\hat{s}_m^{(k)}
=
\begin{cases}
\displaystyle
\frac{s_m^{(k)}}{\sum_{k'\in\mathcal{T}_m}s_m^{(k')}},
& k\in\mathcal{T}_m,\\[10pt]
0,
& \text{otherwise},
\end{cases}
\label{eq:sparse_routing_final}
\end{equation}
where \(\hat{s}_m^{(k)}\) denotes the normalized routing weight of expert \(k\) for semantic block \(m\). The selected expert outputs are then aggregated to form the final block representation:
\begin{equation}
\mathbf{y}_m
=
\sum_{k=1}^{R}
\hat{s}_m^{(k)}\mathbf{y}_m^{(k)}.
\label{eq:expert_agg_final}
\end{equation}

After all semantic blocks are processed, their representations are passed to the prediction head to generate the final forecast:
$
\hat{\mathbf{Y}}
=
\mathrm{Readout}\big(\{\mathbf{y}_m\}_{m=1}^{K_b}\big),
$
where \(\mathrm{Readout}(\cdot)\) denotes the regression head that maps the routed block representations to the forecasting output.

\subsection{Computational Complexity Analysis}

The computational cost of SCPaT mainly arises from semantic representation learning, directed dependency modeling, and expert routing. Let \(N_u=\sum_{c=1}^{C}N_c\) denote the total number of semantic units and \(d_h\) the hidden dimension.
The Semantic Vector Encoder scales approximately linearly with the input size, with complexity \(\mathcal{O}(CLd_h)\) under a fixed number of convolution branches. The Transfer Entropy Graph Constructor evaluates pairwise directed dependencies among semantic units, resulting in a complexity of approximately \(\mathcal{O}(N_u^2d_h^2)\), which constitutes the main quadratic component of SCPaT. Top-\(K_g\) sparsification does not reduce this pairwise computation, but limits the number of retained graph edges and reduces the cost of subsequent graph operations. The routing module operates on \(K_b\) semantic blocks and activates only a subset of the \(R\) experts through Top-\(P\) routing, thereby avoiding the cost of evaluating all experts for every block.

\begin{table*}
\centering
\caption{Summary of the 12 datasets used in our forecasting experiments, including the prediction horizons, data dimensionality, sampling frequency, and total number of time points.}
\label{dataset_detailed}
\scriptsize
\renewcommand{\arraystretch}{1.2}
\begin{tabular}{llcccc}
\toprule
Task Type & Dataset & Prediction Horizons & Time Point & Dimension & Frequency \\
\midrule
\multirow{8}{*}{Long-term Forecasting} 
& ETTh1 & \{96,192,336,720\}&17420 & 7 & Hourly \\
& ETTh2 & \{96,192,336,720\} &17420& 7 & Hourly \\
& ETTm1 & \{96,192,336,720\} &69680& 7 & 15 min \\
& ETTm2 & \{96,192,336,720\} &69680& 7 & 15 min \\
& Weather & \{96,192,336,720\}&52603 & 21 & 10 min \\
& Traffic & \{96,192,336,720\} &17451& 862 & Hourly \\
& Electricity & \{96,192,336,720\}&26211 & 321 & Hourly \\
& Solar & \{96,192,336,720\} &52179& 137 & 10 min \\
\midrule
\multirow{4}{*}{Short-term Forecasting}
& PEMS03 & \{12,24,48\}&26208 & 358 & 5 min \\
& PEMS04 & \{12,24,48\}&16992 & 307 & 5 min \\
& PEMS07 & \{12,24,48\} &28224& 883 & 5 min \\
& PEMS08 & \{12,24,48\} &17856& 170 & 5 min \\
\bottomrule
\end{tabular}
\end{table*}

\section{Experiment}

\subsection{Experimental Setup}
\subsubsection{Datasets}


We conduct experiments on 12 widely used benchmark datasets. For long horizon forecasting, we consider four ETT datasets, including ETTh1, ETTh2, ETTm1, and ETTm2 \citep{zhou2021informer}, which contain electricity transformer temperature and load measurements at different temporal resolutions. We further include Weather \citep{wu2021autoformer}, containing 21 meteorological variables; Traffic, recording road occupancy rates on California freeways; Electricity, containing hourly electricity consumption records from 321 households; and Solar, consisting of solar power production measurements from photovoltaic plants. For short horizon forecasting, we adopt four standard PEMS benchmarks, namely PEMS03, PEMS04, PEMS07, and PEMS08 \citep{zheng2020gman}, which contain traffic flow measurements collected from road sensors. Detailed dataset statistics are summarized in Table~\ref{dataset_detailed}.

\subsubsection{Baselines.}

We compare SCPaT with eight representative baselines covering diverse temporal modeling paradigms and backbone architectures. The Transformer-based methods include PatchTST \citep{patchtst}, iTransformer \citep{itransformer}, DUET \citep{qiu2025duet}, MSPatch \citep{cao2025mspatch}, and Crossformer \citep{crossformer}, which represent different designs for patch modeling, variable interaction, and multiscale temporal representation. We further include TimesNet \citep{wu2022timesnet}, MSGNet \citep{msgnet}, and HDMixer \citep{huang2024hdmixer}, representing CNN, GNN, and MLP architectures, respectively. This selection enables evaluation across heterogeneous modeling paradigms and helps assess whether the improvements of SCPaT generalize beyond a specific backbone family.

\subsubsection{Implementation Details.}


All baseline models and SCPaT are implemented in PyTorch 2.1.2 and evaluated under a unified experimental protocol on a single NVIDIA RTX 4090D GPU with 24 GB of memory. For a fair comparison, we retrain all baselines on the same dataset splits using the recommended hyperparameter settings from their original papers. When such settings are unavailable, hyperparameters are selected through grid search on the validation set.
For long horizon forecasting, we evaluate all models at \(H\in\{96,192,336,720\}\), following the standard experimental protocol. For short horizon forecasting on the PEMS datasets, we use \(H\in\{12,24,48\}\). The hyperparameters of SCPaT are also selected through grid search \citep{pedregosa2011scikit} based on validation performance. Mean squared error (MSE) and mean absolute error (MAE) are used as the evaluation metrics \citep{hyndman2006another}.

\begin{table*}
\renewcommand{\arraystretch}{1.3}
\caption{Multivariate long-term forecasting results over four prediction horizons, $H \in \{96, 192, 336, 720\}$, with the input length fixed at $L=96$. The best and second-best results are marked in \textcolor{red}{\textbf{bold red}} and \textcolor{blue}{\underline{blue underline}}, respectively.}
\resizebox{\textwidth}{!}{%
\begin{tabular}{c|c|cc|cc|cc|cc|cc|cc|cc|cc|cc}
\toprule[1.0pt]
 \multicolumn{2}{c}{\scalebox{1.1}{Models}}& \multicolumn{2}{c}{SCPaT (\textbf{Ours})} & \multicolumn{2}{c}{MSPatch} & \multicolumn{2}{c}{DUET} & \multicolumn{2}{c}{iTransformer} & \multicolumn{2}{c}{MSGNet} & \multicolumn{2}{c}{HDMixer} & \multicolumn{2}{c}{PatchTST} & \multicolumn{2}{c}{TimesNet} & \multicolumn{2}{c}{CrossFormer} \\
 \cmidrule(lr){3-4} \cmidrule(lr){5-6} \cmidrule(lr){7-8} \cmidrule(lr){9-10}
 \cmidrule(lr){11-12} \cmidrule(lr){13-14} \cmidrule(lr){15-16} \cmidrule(lr){17-18} \cmidrule(lr){19-20}

\multicolumn{2}{c}{Metric} & MSE & MAE & MSE & MAE & MSE & MAE & MSE & MAE & MSE & MAE & MSE & MAE & MSE & MAE & MSE & MAE & MSE & MAE \\
\toprule[1.0pt]

\multirow{4}{*}{\rotatebox{90}{ETTh1}}
& 96 & \textcolor{red}{\textbf{0.370}} & \textcolor{red}{\textbf{0.392}} & \textcolor{blue}{\underline{0.372}} & \textcolor{blue}{\underline{0.395}} & 0.384 & 0.402 & 0.386 & 0.405 & 0.390 & 0.411 & 0.386 & 0.401 & 0.414 & 0.419 & 0.384 & 0.402 & 0.423 & 0.448 \\
& 192 & \textcolor{red}{\textbf{0.418}} & \textcolor{red}{\textbf{0.422}} & \textcolor{blue}{\underline{0.428}} & \textcolor{blue}{\underline{0.426}} & 0.437 & 0.429 & 0.441 & 0.436 & 0.442 & 0.442 & 0.441 & 0.430 & 0.460 & 0.445 & 0.446 & 0.439 & 0.471 & 0.474 \\
& 336 & \textcolor{red}{\textbf{0.452}} & \textcolor{red}{\textbf{0.440}} & 0.484 & 0.452 & 0.477 & 0.453 & 0.465 & 0.445 & 0.487 & 0.458 & \textcolor{blue}{\underline{0.458}} & \textcolor{blue}{\underline{0.443}} & 0.501 & 0.466 & 0.491 & 0.469 & 0.570 & 0.546 \\
& 720 & \textcolor{red}{\textbf{0.449}} & \textcolor{red}{\textbf{0.458}} & \textcolor{blue}{\underline{0.480}} & \textcolor{blue}{\underline{0.473}} & 0.486 & 0.475 & 0.503 & 0.491 & 0.494 & 0.488 & 0.512 & 0.484 & 0.500 & 0.488 & 0.521 & 0.500 & 0.653 & 0.621 \\
\midrule

\multirow{4}{*}{\rotatebox{90}{ETTh2}}
& 96 & \textcolor{red}{\textbf{0.288}} & \textcolor{red}{\textbf{0.338}} & \textcolor{blue}{\underline{0.292}} & \textcolor{blue}{\underline{0.344}} & 0.298 & 0.351 & 0.297 & 0.349 & 0.328 & 0.371 & 0.293 & \textcolor{red}{\textbf{0.338}} & 0.302 & 0.348 & 0.340 & 0.374 & 0.745 & 0.584 \\
& 192 & \textcolor{red}{\textbf{0.373}} & \textcolor{blue}{\underline{0.396}} & 0.383 & 0.402 & \textcolor{blue}{\underline{0.374}} & \textcolor{red}{\textbf{0.394}} & 0.380 & 0.400 & 0.402 & 0.414 & 0.379 & \textcolor{blue}{\underline{0.396}} & 0.388 & 0.400 & 0.402 & 0.414 & 0.877 & 0.656 \\
& 336 & \textcolor{red}{\textbf{0.408}} & \textcolor{red}{\textbf{0.427}} & \textcolor{blue}{\underline{0.417}} & \textcolor{blue}{\underline{0.431}} & \textcolor{blue}{\underline{0.417}} & \textcolor{blue}{\underline{0.431}} & 0.428 & 0.432 & 0.435 & 0.443 & 0.426 & 0.444 & 0.426 & 0.433 & 0.452 & 0.452 & 1.043 & 0.731 \\
& 720 & 0.430 & \textcolor{blue}{\underline{0.444}} & \textcolor{red}{\textbf{0.422}} & 0.446 & \textcolor{blue}{\underline{0.427}} & \textcolor{blue}{\underline{0.442}} & 0.430 & 0.445 & 0.441 & 0.444 & 0.429 & 0.447 & 0.431 & 0.446 & 0.462 & 0.468 & 1.104 & 0.763 \\
\midrule

\multirow{4}{*}{\rotatebox{90}{ETTm1}}
& 96 & \textcolor{red}{\textbf{0.320}} & \textcolor{red}{\textbf{0.360}} & 0.324 & 0.363 & 0.329 & \textcolor{blue}{\underline{0.361}} & 0.334 & 0.368 & \textcolor{blue}{\underline{0.323}} & 0.366 & 0.340 & 0.368 & 0.329 & 0.367 & 0.338 & 0.375 & 0.404 & 0.426 \\
& 192 & \textcolor{red}{\textbf{0.362}} & \textcolor{red}{\textbf{0.383}} & 0.367 & \textcolor{blue}{\underline{0.385}} & \textcolor{blue}{\underline{0.364}} & \textcolor{blue}{\underline{0.385}} & 0.377 & 0.391 & 0.376 & 0.397 & 0.382 & 0.386 & 0.367 & \textcolor{blue}{\underline{0.385}} & 0.374 & 0.387 & 0.450 & 0.451 \\
& 336 & \textcolor{red}{\textbf{0.389}} & \textcolor{red}{\textbf{0.402}} & \textcolor{blue}{\underline{0.390}} & \textcolor{blue}{\underline{0.404}} & 0.396 & 0.408 & 0.426 & 0.420 & 0.417 & 0.422 & 0.403 & 0.411 & 0.399 & 0.410 & 0.410 & 0.411 & 0.532 & 0.515 \\
& 720 & \textcolor{blue}{\underline{0.456}} & \textcolor{red}{\textbf{0.436}} & \textcolor{red}{\textbf{0.454}} & \textcolor{blue}{\underline{0.437}} & 0.469 & \textcolor{blue}{\underline{0.437}} & 0.491 & 0.459 & 0.481 & 0.458 & 0.473 & 0.439 & \textcolor{red}{\textbf{0.454}} & 0.439 & 0.478 & 0.450 & 0.666 & 0.589 \\
\midrule

\multirow{4}{*}{\rotatebox{90}{ETTm2}}
& 96 & \textcolor{red}{\textbf{0.173}} & \textcolor{red}{\textbf{0.258}} & \textcolor{blue}{\underline{0.175}} & 0.260 & 0.179 & 0.262 & 0.180 & 0.264 & 0.181 & 0.262 & 0.183 & 0.266 & 0.181 & \textcolor{blue}{\underline{0.259}} & 0.187 & 0.267 & 0.287 & 0.366 \\
& 192 & \textcolor{red}{\textbf{0.238}} & \textcolor{red}{\textbf{0.300}} & \textcolor{blue}{\underline{0.239}} & \textcolor{blue}{\underline{0.302}} & 0.240 & \textcolor{red}{\textbf{0.300}} & 0.250 & 0.309 & 0.247 & 0.307 & 0.247 & 0.307 & 0.241 & \textcolor{blue}{\underline{0.302}} & 0.249 & 0.309 & 0.414 & 0.492 \\
& 336 & \textcolor{red}{\textbf{0.297}} & \textcolor{red}{\textbf{0.339}} & \textcolor{red}{\textbf{0.297}} & \textcolor{blue}{\underline{0.340}} & \textcolor{blue}{\underline{0.302}} & 0.341 & 0.311 & 0.348 & 0.312 & 0.346 & 0.305 & 0.345 & 0.305 & 0.343 & 0.321 & 0.351 & 0.597 & 0.543 \\
& 720 & \textcolor{red}{\textbf{0.394}} & \textcolor{red}{\textbf{0.394}} & \textcolor{blue}{\underline{0.396}} & \textcolor{blue}{\underline{0.397}} & 0.399 & \textcolor{blue}{\underline{0.397}} & 0.412 & 0.407 & 0.414 & 0.403 & 0.406 & 0.400 & 0.402 & 0.406 & 0.408 & 0.403 & 1.730 & 1.042 \\
\midrule

\multirow{4}{*}{\rotatebox{90}{Weather}}
& 96 & \textcolor{red}{\textbf{0.163}} & \textcolor{red}{\textbf{0.208}} & \textcolor{blue}{\underline{0.165}} & \textcolor{blue}{\underline{0.209}} & 0.167 & \textcolor{blue}{\underline{0.209}} & 0.174 & 0.214 & 0.169 & 0.212 & 0.174 & 0.223 & 0.177 & 0.218 & 0.172 & 0.220 & 0.171 & 0.230 \\
& 192 & \textcolor{red}{\textbf{0.209}} & \textcolor{red}{\textbf{0.250}} & \textcolor{blue}{\underline{0.211}} & \textcolor{blue}{\underline{0.252}} & 0.212 & 0.254 & 0.221 & 0.254 & 0.218 & 0.255 & 0.225 & 0.264 & 0.225 & 0.259 & 0.219 & 0.261 & 0.226 & 0.277 \\
& 336 & \textcolor{red}{\textbf{0.266}} & \textcolor{red}{\textbf{ 0.292}} &\textcolor{blue}{\underline{ 0.268}} &\textcolor{blue}{\underline{0.294}} & 0.269 & 0.297 & 0.278 & 0.296 & 0.272 & 0.299 & 0.277 & 0.301 & 0.278 & 0.297 & 0.280 & 0.306 & 0.272 & 0.335 \\
& 720 & \textcolor{blue}{\underline{0.345}} & 0.349 & \textcolor{red}{\textbf{0.342}} & \textcolor{red}{\textbf{0.345}} & 0.348 & \textcolor{blue}{\underline{0.347}} & 0.358 & 0.349 & 0.350 & 0.348 & 0.349 & \textcolor{blue}{\underline{0.347}} & 0.354 & 0.348 & 0.365 & 0.359 & 0.398 & 0.418 \\
\midrule

\multirow{4}{*}{\rotatebox{90}{Traffic}}
& 96 & \textcolor{red}{\textbf{0.393}} & \textcolor{red}{\textbf{0.247}} & 0.460 & 0.295 & \textcolor{blue}{\underline{0.395}} & \textcolor{blue}{\underline{0.256}} & \textcolor{blue}{\underline{0.395}} & 0.268 & 0.605 & 0.344 & 0.529 & 0.353 & 0.544 & 0.359 & 0.593 & 0.321 & 0.522 & 0.290 \\
& 192 & 0.425 & \textcolor{red}{\textbf{0.259}} & 0.466 & 0.300 & \textcolor{blue}{\underline{0.420}} & \textcolor{blue}{\underline{0.266}} & \textcolor{red}{\textbf{0.417}} & 0.276 & 0.613 & 0.359 & 0.535 & 0.361 & 0.540 & 0.354 & 0.617 & 0.336 & 0.530 & 0.293 \\
& 336 & \textcolor{red}{\textbf{0.454}} & \textcolor{red}{\textbf{0.270}} & 0.484 & 0.306 & \textcolor{blue}{\underline{0.458}} & \textcolor{blue}{\underline{0.272}} & 0.461 & 0.283 & 0.642 & 0.376 & 0.541 & 0.361 & 0.551 & 0.358 & 0.629 & 0.336 & 0.558 & 0.305 \\
& 720 & \textcolor{red}{\textbf{0.489}} & \textcolor{blue}{\underline{0.291}} & 0.510 & 0.323 & 0.499 & \textcolor{red}{\textbf{0.288}} & \textcolor{blue}{\underline{0.497}} & 0.302 & 0.702 & 0.401 & 0.591 & 0.389 & 0.586 & 0.375 & 0.640 & 0.350 & 0.589 & 0.328 \\
\midrule

\multirow{4}{*}{\rotatebox{90}{Electricity}}
& 96 & \textcolor{red}{\textbf{0.139}} & \textcolor{red}{\textbf{0.233}} & 0.158 & 0.257 & \textcolor{blue}{\underline{0.148}} & \textcolor{blue}{\underline{0.236}} & 0.149 & 0.240 & 0.165 & 0.274 & 0.163 & 0.275 & 0.195 & 0.285 & 0.168 & 0.272 & 0.219 & 0.314 \\
& 192 & \textcolor{red}{\textbf{0.155}} & \textcolor{red}{\textbf{0.248}} & 0.171 & 0.268 & 0.164 & \textcolor{blue}{\underline{0.249}} & \textcolor{blue}{\underline{0.162}} & 0.253 & 0.184 & 0.292 & 0.186 & 0.281 & 0.195 & 0.285 & 0.184 & 0.289 & 0.231 & 0.322 \\
& 336 & \textcolor{red}{\textbf{0.168}} & \textcolor{red}{\textbf{0.263}} & 0.184 & 0.283 & \textcolor{blue}{\underline{0.178}} & \textcolor{blue}{\underline{0.265}} & \textcolor{blue}{\underline{0.178}} & 0.269 & 0.195 & 0.302 & 0.199 & 0.295 & 0.215 & 0.305 & 0.198 & 0.300 & 0.246 & 0.337 \\
& 720 & \textcolor{red}{\textbf{0.200}} & \textcolor{red}{\textbf{0.293}} & 0.223 & 0.308 & \textcolor{blue}{\underline{0.206}} & \textcolor{blue}{\underline{0.303}} & 0.225 & 0.317 & 0.231 & 0.332 & 0.231 & 0.321 & 0.256 & 0.337 & 0.220 & 0.320 & 0.280 & 0.363 \\
\midrule

\multirow{4}{*}{\rotatebox{90}{Solar}}
& 96  &\textcolor{red}{\textbf{ 0.202}} & \textcolor{blue}{\underline{0.234}} &\textcolor{blue}{\underline{ 0.204}} & 0.237 & 0.205 & \textcolor{red}{\textbf{0.217}} & 0.203 & 0.237 & 0.208 & 0.243 & 0.207 &0.244 & 0.234 & 0.286 & 0.250 & 0.292 & 0.310 & 0.331 \\
& 192 & \textcolor{red}{\textbf{0.228}} & \textcolor{blue}{\underline{0.258}} & \textcolor{blue}{\underline{0.231}} & 0.260 & \textcolor{blue}{\underline{0.231}} & \textcolor{red}{\textbf{0.239}} & 0.233 & 0.261 & 0.258 & 0.281 & 0.245 & 0.273 & 0.267 & 0.310 & 0.296 & 0.318 & 0.734 & 0.725 \\
& 336 & \textcolor{red}{\textbf{0.239}} & \textcolor{blue}{\underline{ 0.270}} & \textcolor{blue}{\underline{0.240}} & 0.273 & 0.243 & \textcolor{red}{\textbf{0.242}} & 0.248 & 0.273 & 0.293 & 0.311 & 0.263 & 0.285 & 0.290 & 0.315 & 0.319 & 0.330 & 0.750 & 0.735 \\
& 720 &  \textcolor{red}{\textbf{0.250}} & 0.278 & \textcolor{red}{\textbf{0.250}} & \textcolor{blue}{\underline{0.275}} & 0.255 & \textcolor{red}{\textbf{0.256}} & 0.260 & 0.280 & 0.290 & 0.315 & 0.288 & 0.299 & 0.289 & 0.317 & 0.338 & 0.337 & 0.769 & 0.765 \\
\midrule[0.5pt]

\rowcolor{gray!20}
\multicolumn{2}{c|}{\(\mathrm{1^{st}}\) Count} &
\textcolor{red}{\textbf{28}} & \textcolor{red}{\textbf{24}} &
\textcolor{blue}{\underline{5}} &1 &
0 & \textcolor{blue}{\underline{7}} &
1 & 0 &
0 & 0 &
0 & 1 &
1 & 0 &
0 & 0 &
0 & 0 \\
\bottomrule[1.0pt]
\end{tabular}
}
\label{full_long_result}
\end{table*}

\subsection{Main Results}

\subsubsection{Long-term Forecasting.}

Table~\ref{full_long_result} summarizes the long-term forecasting results. SCPaT achieves the best or second-best performance in most settings, showing consistent effectiveness across different datasets and forecasting horizons. Averaged over all forecasting horizons on the four ETT benchmarks, SCPaT reduces the MSE by 4.6\%, 7.1\%, and 4.1\% compared with PatchTST, TimesNet, and HDMixer, respectively. It also consistently outperforms Crossformer across the evaluated horizons.
These improvements are observed against baselines with different temporal modeling strategies. PatchTST relies on fixed length patching, while TimesNet and HDMixer capture temporal variation through multi period representations and hierarchical mixing, respectively. In contrast, SCPaT adaptively constructs semantic units according to local temporal characteristics and further models dependencies among the resulting units. These results indicate that SCPaT consistently benefits long horizon forecasting across different datasets and prediction lengths, supporting the effectiveness of its adaptive semantic partitioning and dependency modeling.

\begin{table*}
\caption{Multivariate short-term forecasting results over three prediction horizons, $H \in \{12, 24, 48\}$, with the input length fixed at $L=96$. The best and second-best results are marked in \textcolor{red}{\textbf{bold red}} and \textcolor{blue}{\underline{blue underline}}, respectively.}
\renewcommand{\arraystretch}{1.3}
\centering
\resizebox{\textwidth}{!}{%
\begin{tabular}{c|c|cc|cc|cc|cc|cc|cc|cc|cc|cc}
\toprule[1.0pt]
 \multicolumn{2}{c}{\scalebox{1.1}{Models}}& \multicolumn{2}{c}{SCPaT (\textbf{Ours})}& \multicolumn{2}{c}{MSPatch}  & \multicolumn{2}{c}{DUET}& \multicolumn{2}{c}{iTransformer} & \multicolumn{2}{c}{MSGNet} & \multicolumn{2}{c}{HDMixer}& \multicolumn{2}{c}{TimesNet}& \multicolumn{2}{c}{PatchTST} & \multicolumn{2}{c}{CrossFormer}  \\

 \cmidrule(lr){3-4} \cmidrule(lr){5-6} \cmidrule(lr){7-8} \cmidrule(lr){9-10}
 \cmidrule(lr){11-12} \cmidrule(lr){13-14} \cmidrule(lr){15-16} \cmidrule(lr){17-18}
 \cmidrule(lr){19-20}

\multicolumn{2}{c}{Metric} & MSE & MAE& MSE & MAE & MSE & MAE & MSE & MAE & MSE & MAE & MSE & MAE & MSE & MAE & MSE & MAE & MSE & MAE \\
\toprule[1.0pt]

\multirow{3}{*}{PEMS03}
& 12 & \textcolor{red}{\textbf{0.066}} & \textcolor{red}{\textbf{0.169}} &0.074&0.179& \textcolor{blue}{\underline{0.071}} & 0.176  & \textcolor{blue}{\underline{0.071}} & \textcolor{blue}{\underline{0.174}} & 0.083 & 0.197&0.076 &0.184 &0.085 &0.192 & 0.099 & 0.216& 0.090 & 0.203   \\

& 24 & \textcolor{blue}{\underline{0.086}} & \textcolor{red}{\textbf{0.195}}&0.097&0.200  & \textcolor{red}{\textbf{0.089 }}& \textcolor{blue}{\underline{0.197}}  & 0.093 & 0.201 & 0.097 & 0.217&0.095 &0.210 &0.118 &0.223 & 0.142 & 0.259 & 0.121 & 0.240  \\

& 48 &\textcolor{blue}{\underline{ 0.122}} & \textcolor{blue}{\underline{0.235}}&0.131&0.244  & \textcolor{red}{\textbf{0.118}} & \textcolor{red}{\textbf{0.229}}  & 0.125 & 0.236 & 0.135 & 0.241&0.122 &0.245 &0.155 &0.260& 0.211 & 0.319  & 0.202 & 0.317  \\
\midrule

\multirow{3}{*}{PEMS04}
& 12 & \textcolor{red}{\textbf{0.073}} & \textcolor{red}{\textbf{0.176}}&\textcolor{blue}{\underline{0.075}}&0.185 & 0.076 & \textcolor{blue}{\underline{0.179}}  & 0.078 & 0.183 & 0.082 & 0.187&0.084 &0.188 &0.087 &0.195& 0.105 & 0.224& 0.098 & 0.218 \\

& 24 & \textcolor{red}{\textbf{0.091}} & \textcolor{red}{\textbf{0.199}}&\textcolor{blue}{\underline{0.093}}&\textcolor{blue}{\underline{0.201}}  & 0.097 & 0.203  & 0.095 & 0.205 & 0.128 & 0.241&0.097 &0.210 &0.103 &0.215  & 0.153 & 0.257& 0.131 & 0.256   \\

& 48 & \textcolor{blue}{\underline{0.124}} & \textcolor{red}{\textbf{0.237}}&0.134&0.244  & \textcolor{red}{\textbf{0.119}} &\textcolor{blue}{\underline{ 0.239}}  & 0.126 & 0.240 & 0.192 & 0.315&0.134 &0.253 &0.136 &0.250& 0.229 & 0.339& 0.205 & 0.326   \\
\midrule

\multirow{3}{*}{PEMS07}
& 12 & \textcolor{red}{\textbf{0.058}} & \textcolor{red}{\textbf{0.153}}&0.066&\textcolor{blue}{\underline{0.162}} & \textcolor{blue}{\underline{0.059}} & 0.166  & 0.067 & 0.165 & 0.087 & 0.191 &0.074 &0.182 &0.082 &0.181& 0.095 & 0.207& 0.094 & 0.200\\

& 24 & \textcolor{red}{\textbf{0.074}} & \textcolor{red}{\textbf{0.174}}&0.083&0.189  & \textcolor{blue}{\underline{0.079}} &\textcolor{blue}{\underline{ 0.183}} & 0.088 & 0.190 & 0.135 & 0.241&0.089 &0.197 &0.101 &0.204 & 0.150 & 0.262& 0.139 & 0.247   \\

& 48 & \textcolor{red}{\textbf{0.102}} & \textcolor{red}{\textbf{0.207}}&0.112&0.217 & \textcolor{blue}{\underline{0.106}} & \textcolor{blue}{\underline{0.211}}  & 0.110 & 0.215 & 0.297 & 0.351&0.124 &0.217 &0.134 &0.238& 0.253 & 0.340& 0.311 & 0.369   \\
\midrule

\multirow{3}{*}{PEMS08}
& 12 & \textcolor{red}{\textbf{0.070}} & \textcolor{red}{\textbf{0.173}}&0.079&\textcolor{blue}{\underline{0.175}}  & \textcolor{blue}{\underline{0.077}} & 0.177 & 0.079 & 0.182 & 0.159 & 0.208&0.098 &0.193 &0.112 &0.212& 0.168 & 0.232& 0.165 & 0.214   \\

& 24 & \textcolor{red}{\textbf{0.085}} & \textcolor{red}{\textbf{0.187}}&\textcolor{blue}{\underline{0.094}}&0.211  & 0.098 & \textcolor{blue}{\underline{0.201}}  & 0.115 & 0.219 & 0.210 & 0.251&0.108 &0.223 &0.141 &0.238 & 0.224 & 0.281& 0.215 & 0.260   \\

& 48 & \textcolor{red}{\textbf{0.112}} & \textcolor{red}{\textbf{0.220}}&0.150&0.232  & \textcolor{blue}{\underline{0.139}} & \textcolor{blue}{\underline{0.227}}  & 0.186 & 0.235 & 0.311 & 0.358&0.167 &0.247 &0.198 &0.283& 0.321 & 0.354 & 0.315 & 0.335  \\
\midrule[0.5pt]

\rowcolor{gray!20}
\multicolumn{2}{c|}{\(\mathrm{1^{st}}\) Count} &
\textcolor{red}{\textbf{9}} & \textcolor{red}{\textbf{11}}&0&0 & \textcolor{blue}{\underline{3}} & \textcolor{blue}{\underline{1}} & 0 & 0 & 0 & 0 & 0 & 0 & 0 & 0 & 0 & 0 & 0 & 0  \\

\bottomrule[1.0pt]
\end{tabular}
}
\label{full_short_result}
\end{table*}


\subsubsection{Short-term Forecasting.}

Table~\ref{full_short_result} reports the short-term forecasting results on the four PEMS benchmarks. SCPaT ranks first in 21 of the 24 evaluated cases and second in the remaining three, showing consistent performance across datasets and forecasting horizons. Averaged over all settings, SCPaT achieves an MSE of 0.0886 and an MAE of 0.1938, reducing the corresponding errors of DUET, the strongest baseline on average, by 5.8\% and 2.6\%, respectively. The improvement is particularly pronounced on PEMS08, where SCPaT reduces the MSE from 0.139 to 0.112 at \(H=48\), corresponding to a 19.4\% improvement. It also achieves the lowest MSE and MAE across all three horizons on both PEMS07 and PEMS08.
Despite the relatively short forecasting horizons, the PEMS datasets contain complex local dynamics characterized by recurring patterns, rapid fluctuations, and abrupt changes. The consistent improvements across datasets and horizons indicate that SCPaT remains effective under such short horizon forecasting settings. In particular, the strong performance on PEMS07 and PEMS08 supports the effectiveness of adaptive semantic partitioning for modeling complex traffic dynamics.

\subsection{Model Analysis}

\subsubsection{Routing Analysis.}

We analyze the behavior of the Importance Aware Router by averaging the expert assignment probabilities over all semantic blocks in the test set, as summarized in Table~\ref{tab:routing}. The results exhibit distinct routing patterns across datasets. For ETTh1 and ETTm1, the probabilities are relatively distributed across the three experts, with average assignments of 0.62/0.21/0.17 and 0.51/0.25/0.24, respectively. In contrast, Weather exhibits a more concentrated routing pattern, with Expert 1 receiving an average probability of 0.81 and substantially smaller probabilities assigned to the other experts. These differences show that the router adjusts its expert allocation according to the characteristics of the input semantic blocks rather than applying a fixed assignment pattern across datasets.

\begin{figure}
 \centering
\includegraphics[width=0.98\columnwidth]{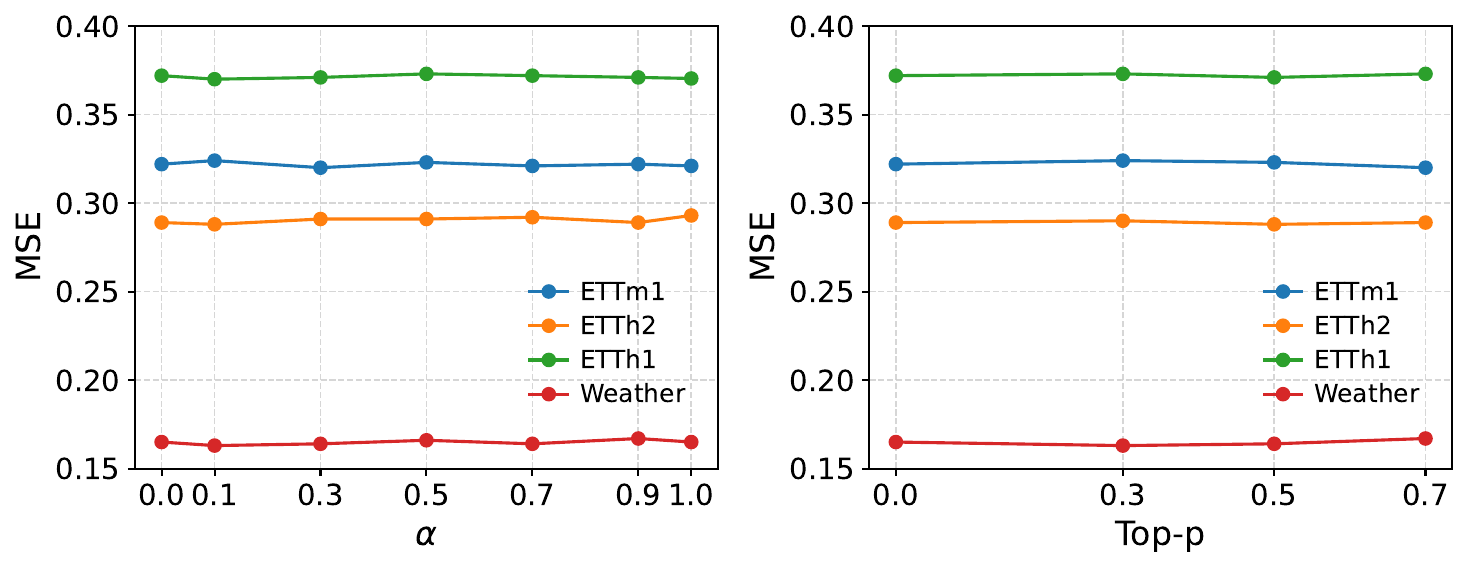}
\caption{Sensitivity analysis of the hyperparameters $\alpha$ and Top-P on ETTm1, ETTh1, ETTh2, and Weather, with both the input length and prediction horizon fixed at 96.}
\label{fig3}
\end{figure}

\begin{table}
\centering
\caption{Average routing probability distribution across experts, with both the input length and prediction horizon fixed at 96.}
\label{tab:routing}
\setlength{\tabcolsep}{3pt}
\renewcommand{\arraystretch}{1}
\scriptsize
\begin{tabular*}{\columnwidth}{@{\extracolsep{\fill}} l ccc}
\toprule
Dataset & Expert 1 & Expert 2 & Expert 3 \\
\midrule
ETTh1   & 0.62 & 0.21 & 0.17 \\
ETTm1   & 0.51 & 0.25 & 0.24 \\
Weather & 0.81 & 0.11 & 0.08 \\
\bottomrule
\end{tabular*}
\end{table}

\subsubsection{Hyperparameter Sensitivity}

We further investigate the sensitivity of SCPaT to the graph sparsification ratio \(\alpha\) and the Top \(P\) routing threshold, as shown in Figure~\ref{fig3}. SCPaT maintains relatively stable performance across a broad range of parameter values, indicating limited sensitivity to either hyperparameter.
For the sparsification ratio \(\alpha\), the preferred setting varies across datasets. ETTh1, ETTh2, and Weather achieve their best performance at \(\alpha=0.1\), whereas ETTm1 favors a larger value of \(\alpha=0.3\). This variation suggests that different datasets may require different levels of graph connectivity to preserve informative dependencies while suppressing marginal interactions.
A similar dataset dependent pattern is observed for the Top \(P\) threshold. ETTh1 performs best at \(P=0.5\), Weather at \(P=0.3\), and ETTm1 around \(P=0.7\). A smaller \(P\) results in more selective expert activation, whereas a larger \(P\) retains more experts and reduces routing sparsity. The results indicate that moderate sparsity generally provides a favorable balance between selective routing and sufficient expert coverage.

\subsubsection{Robustness to Missing and Noisy Observations.}

We first evaluate robustness under incomplete observations by randomly masking input values at missing rates ranging from 0\% to 30\%. The masked entries are replaced with zeros, while the remaining observations are left unchanged. As shown in Table~\ref{robustness}, the forecasting errors of all methods increase as the missing rate grows. Nevertheless, SCPaT consistently achieves lower errors than the compared baselines across different missing rates, indicating stronger robustness to incomplete input observations.

\begin{figure}
 \centering
\includegraphics[width=0.98\columnwidth]{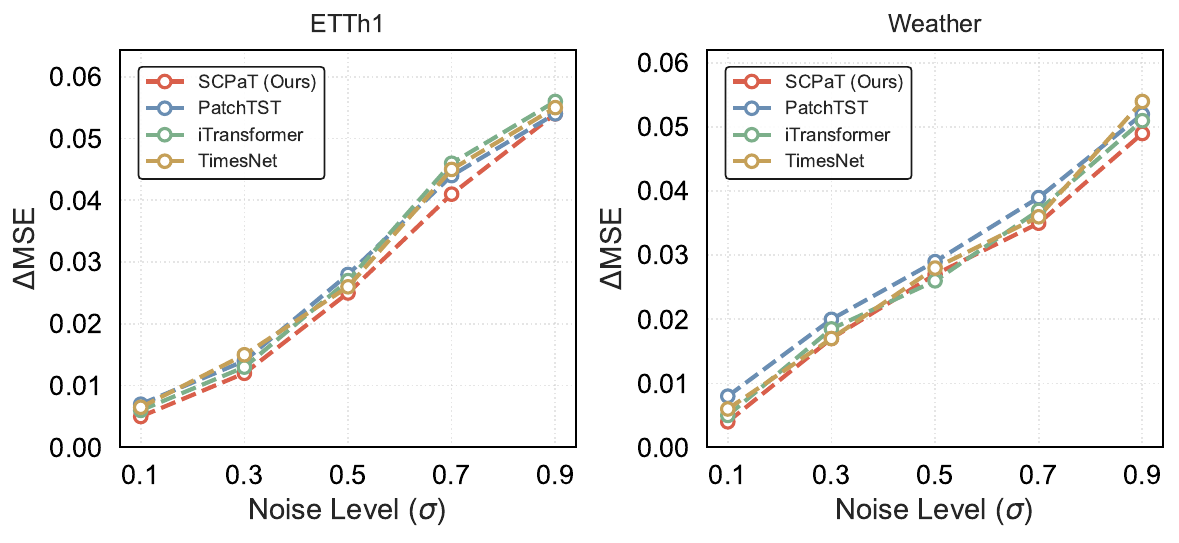}
\caption{Performance degradation under different Gaussian noise levels on ETTh1 and Weather with both the input length and prediction horizon fixed at 96.}
\label{fignew}
\end{figure}

We further evaluate robustness to input perturbations by adding zero mean Gaussian noise to the test sequences with standard deviation \(\sigma \in \{0.1,0.3,0.5,0.7,0.9\}\). Figure~\ref{fignew} compares SCPaT with PatchTST, iTransformer, and TimesNet under increasing noise levels. Although the MSE of all methods increases with \(\sigma\), SCPaT exhibits a consistently smaller performance degradation across the evaluated noise levels. These results indicate that SCPaT maintains relatively stable forecasting performance under both missing observations and additive input noise.

\subsubsection{Look-back Window.}

\begin{figure*}
\centering
\includegraphics[width=0.98\textwidth]{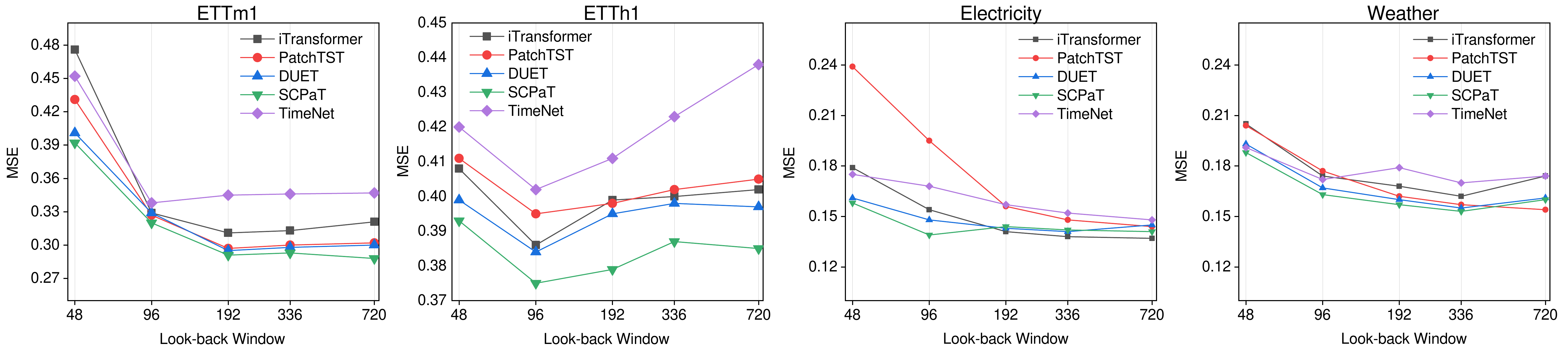}
\caption{Performance comparison under different look-back window lengths on ETTm1, ETTh1, Electricity, and Weather, with $L \in \{48, 96, 192, 336, 720\}$ and the prediction horizon fixed at 96.}
\label{fig4}
\end{figure*}

Longer input sequences can provide richer historical context for forecasting, but existing models do not necessarily benefit from increasing the look-back window, and their performance may vary across different input lengths~\citep{shao2024exploring}. As shown in Figure~\ref{fig4}, extending the look-back window does not consistently improve forecasting accuracy, suggesting that additional historical observations may also introduce redundant or less informative patterns. In contrast, SCPaT maintains consistently competitive performance across different look-back window settings, with relatively small variations as the input length changes. This behavior indicates that SCPaT is less sensitive to the choice of historical window and can make effective use of temporal information under different input configurations.

\begin{table*}
\centering
\caption{Robustness analysis on the ETTm2 and ETTh1 datasets under different missing rates, with the input length fixed at $L=96$ and the prediction horizon fixed at $H=96$. The best results are highlighted in bold.}
\label{robustness}
\setlength{\tabcolsep}{6pt}
\renewcommand{\arraystretch}{1.2}
\scriptsize
\begin{tabular}{c|cccccc|cccccc}
\toprule
\multirow{4}{*}{Missing Rate} 
& \multicolumn{6}{c|}{ETTm2} 
& \multicolumn{6}{c}{ETTh1} \\
\cmidrule(lr){2-7} \cmidrule(lr){8-13}
& \multicolumn{2}{c|}{SCPaT (Ours)} 
& \multicolumn{2}{c|}{iTransformer} 
& \multicolumn{2}{c|}{PatchTST}
& \multicolumn{2}{c|}{SCPaT (Ours)} 
& \multicolumn{2}{c|}{iTransformer} 
& \multicolumn{2}{c}{PatchTST} \\
\cmidrule(lr){2-3} \cmidrule(lr){4-5} \cmidrule(lr){6-7}
\cmidrule(lr){8-9} \cmidrule(lr){10-11} \cmidrule(lr){12-13}
& MSE & MAE & MSE & MAE & MSE & MAE 
& MSE & MAE & MSE & MAE & MSE & MAE \\
\midrule
0.00 & \textbf{0.173} & \textbf{0.258} & 0.180 & 0.264 & 0.181 & 0.259 
     & \textbf{0.370} & \textbf{0.392} & 0.386 & 0.405 & 0.414 & 0.419 \\
0.05 & \textbf{0.213} & \textbf{0.297} & 0.249 & 0.441 & 0.252 & 0.327 
     & \textbf{0.379} & \textbf{0.401} & 0.404 & 0.413 & 0.436 & 0.428 \\
0.10 & \textbf{0.248} & \textbf{0.326} & 0.295 & 0.472 & 0.328 & 0.376 
     & \textbf{0.390} & \textbf{0.412} & 0.433 & 0.424 & 0.455 & 0.441 \\
0.15 & \textbf{0.287} & \textbf{0.354} & 0.379 & 0.559 & 0.411 & 0.423 
     & \textbf{0.404} & \textbf{0.424} & 0.462 & 0.432 & 0.467 & 0.460 \\
0.20 & \textbf{0.332} & \textbf{0.385} & 0.433 & 0.567 & 0.502 & 0.471 
     & \textbf{0.419} & \textbf{0.438} & 0.478 & 0.448 & 0.482 & 0.483 \\
0.25 & \textbf{0.388} & \textbf{0.419} & 0.506 & 0.574 & 0.601 & 0.520 
     & \textbf{0.437} & \textbf{0.454} & 0.487 & 0.473 & 0.498 & 0.503 \\
0.30 & \textbf{0.452} & \textbf{0.456} & 0.596 & 0.580 & 0.709 & 0.569 
     & \textbf{0.459} & \textbf{0.471} & 0.500 & 0.498 & 0.511 & 0.527 \\
\bottomrule
\end{tabular}
\end{table*}

\begin{figure}
\centering
\includegraphics[width=0.98\columnwidth]{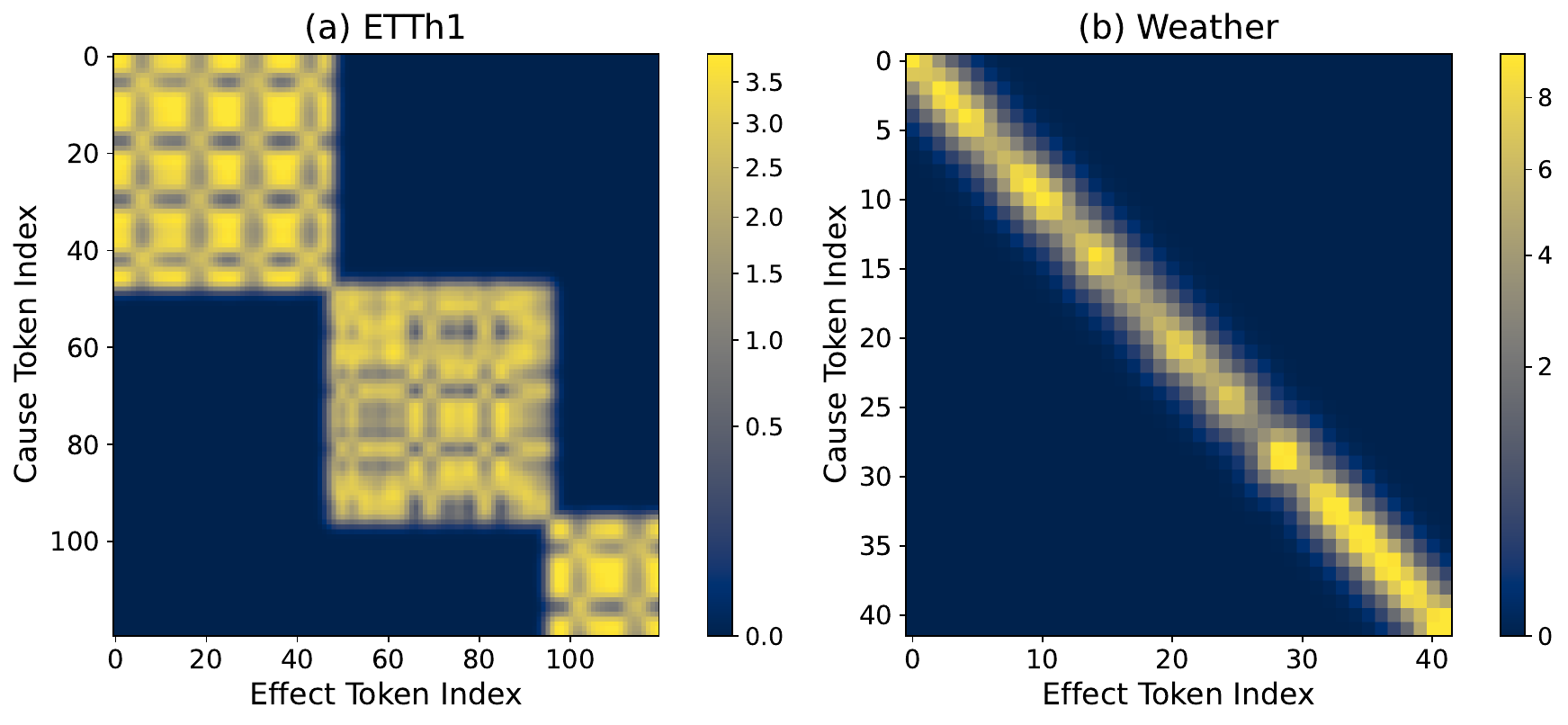}
\caption{Heatmaps of the learned directed adjacency matrix $A_{\text{final}}$ on the ETTh1 and Weather datasets.}
\label{fig5}
\end{figure}

\begin{figure*}
    \centering
    \newcommand{\imgwidth}{0.23\textwidth}
    \setlength{\tabcolsep}{3pt}
    \begin{tabular}{@{}c cccc@{}}
        \rotatebox{90}{\bf Traffic} &
        \begin{minipage}{\imgwidth}
            \includegraphics[width=\linewidth]{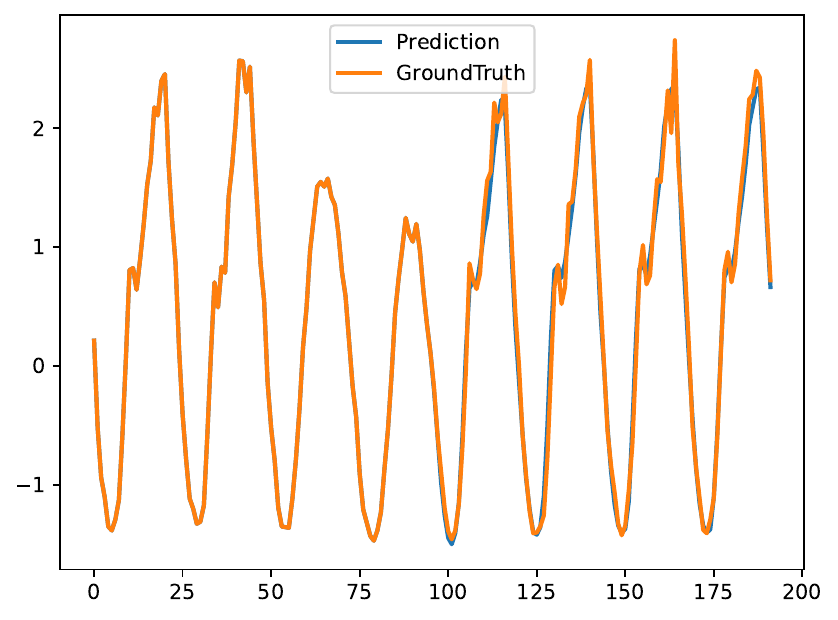}
            \centering
        \end{minipage} &
        \begin{minipage}{\imgwidth}
            \includegraphics[width=\linewidth]{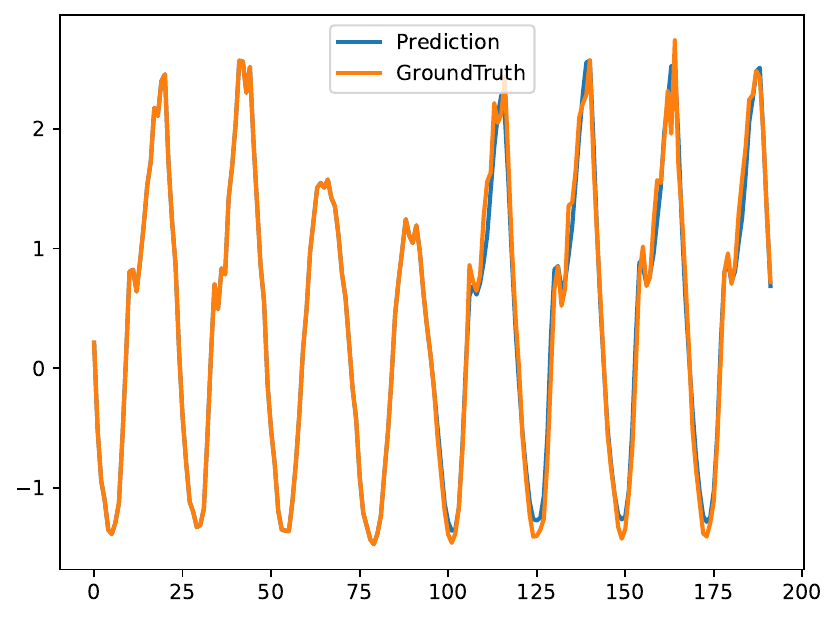}
            \centering
        \end{minipage} &
        \begin{minipage}{\imgwidth}
            \includegraphics[width=\linewidth]{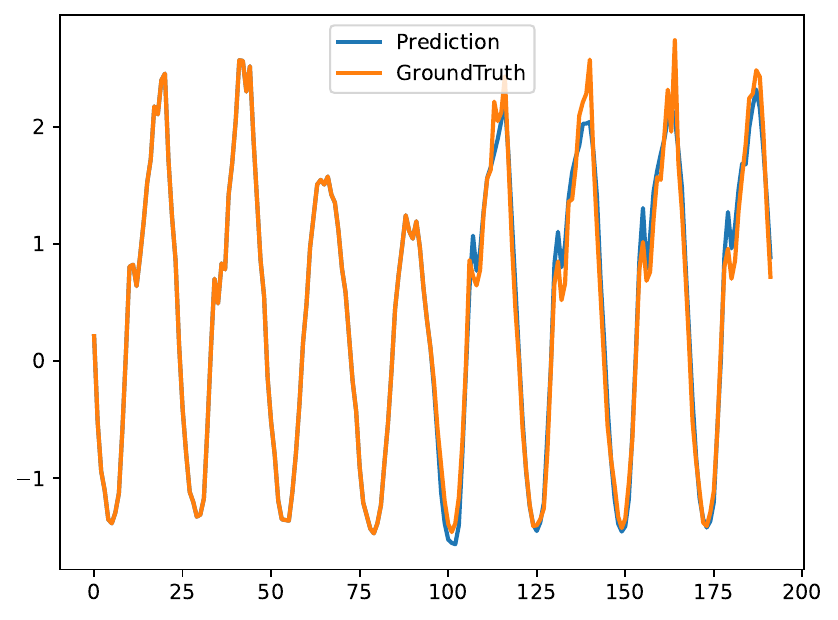}
            \centering
        \end{minipage} &
        \begin{minipage}{\imgwidth}
            \includegraphics[width=\linewidth]{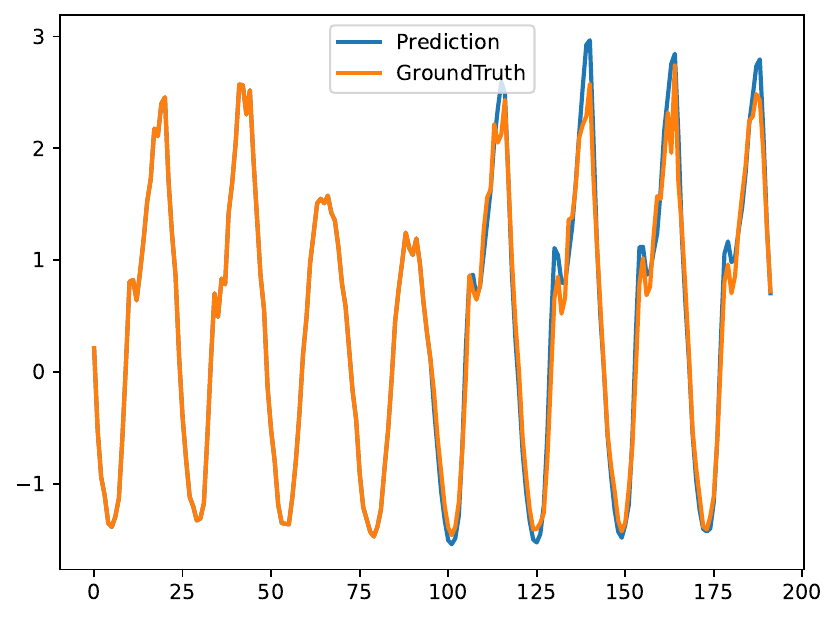}
            \centering
        \end{minipage} \\[0.2em]
        
        \rotatebox{90}{\bf Electricity} &
        \begin{minipage}{\imgwidth}
            \includegraphics[width=\linewidth]{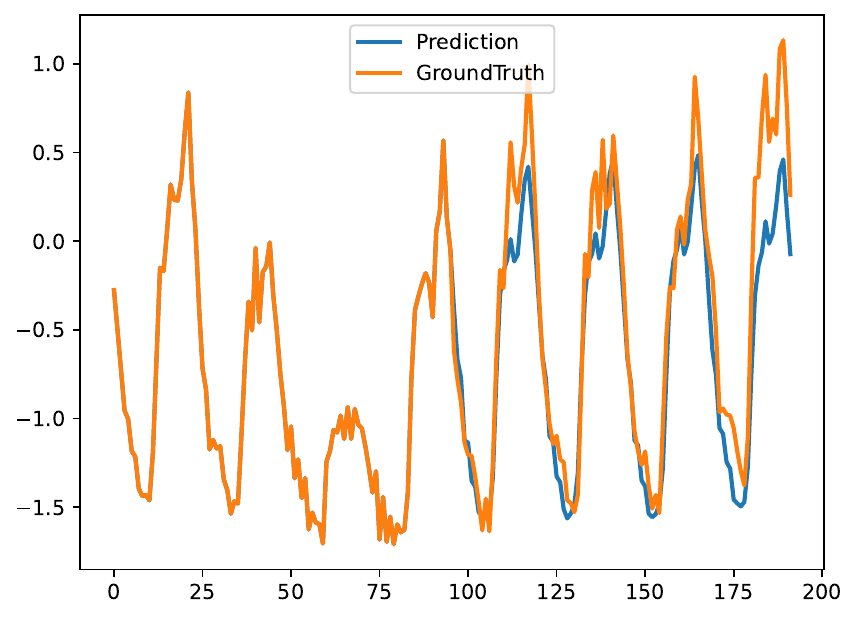}
            \centering
        \end{minipage} &
        \begin{minipage}{\imgwidth}
            \includegraphics[width=\linewidth]{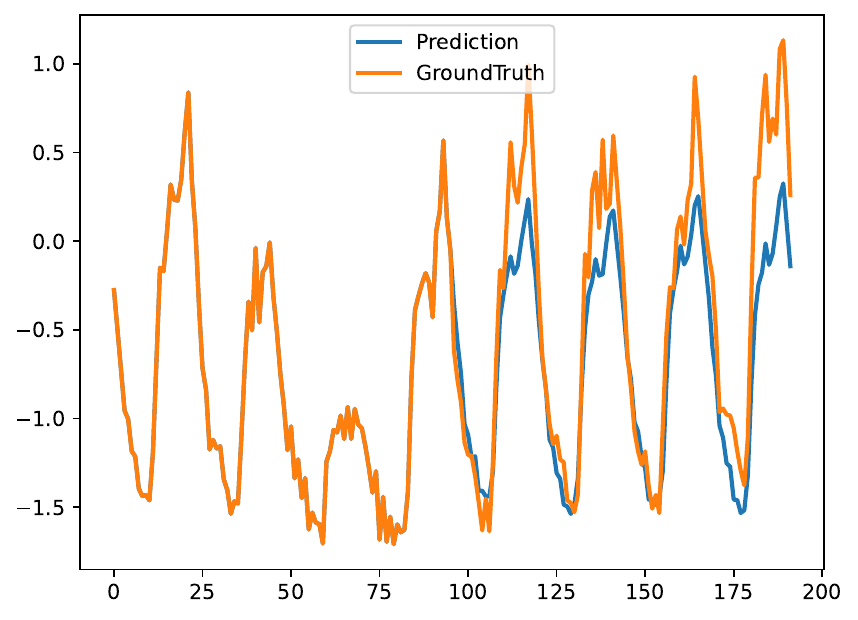}
            \centering
        \end{minipage} &
        \begin{minipage}{\imgwidth}
            \includegraphics[width=\linewidth]{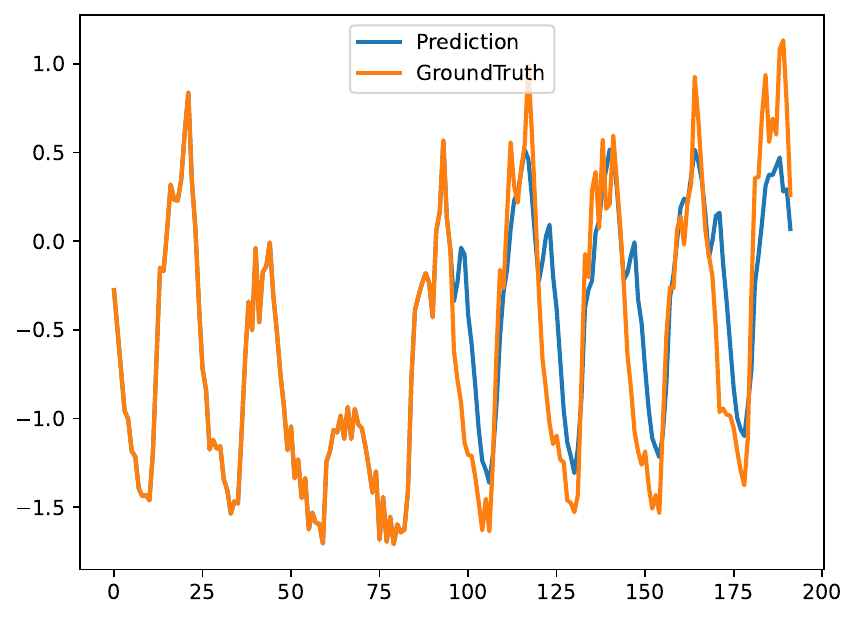}
            \centering
        \end{minipage} &
        \begin{minipage}{\imgwidth}
            \includegraphics[width=\linewidth]{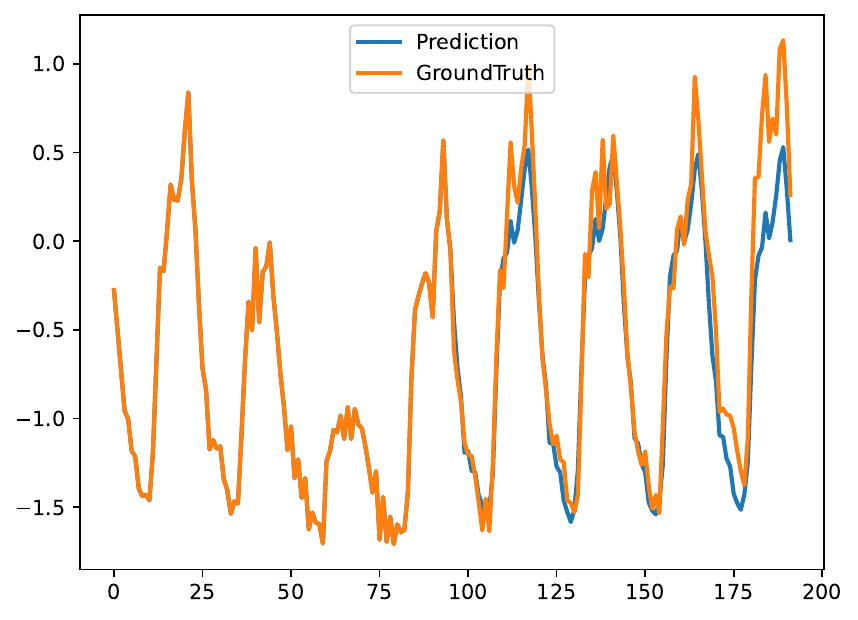}
            \centering
        \end{minipage} \\[0.2em]
        
        \rotatebox{90}{\bf ETTh2} &
        \begin{minipage}{\imgwidth}
            \includegraphics[width=\linewidth]{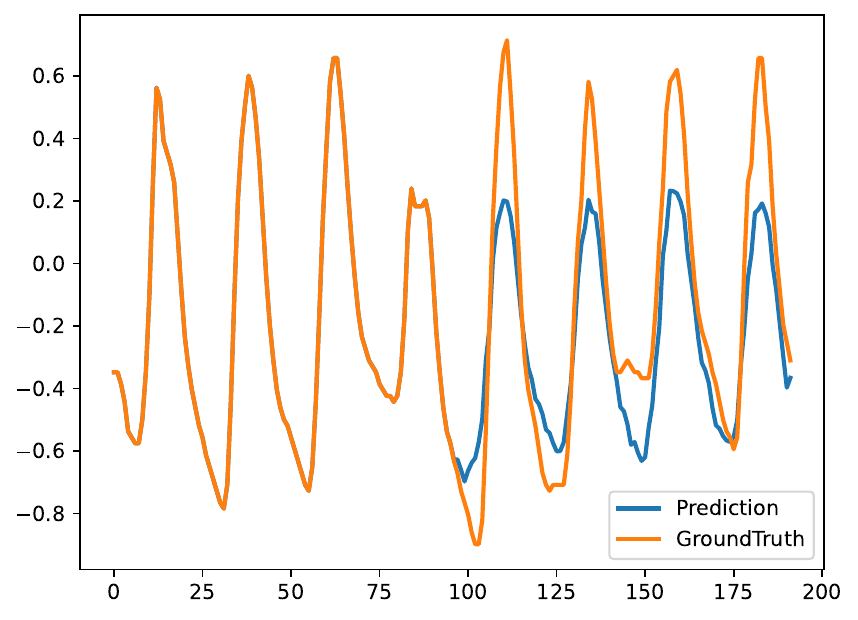}
            \centering
        \end{minipage} &
        \begin{minipage}{\imgwidth}
            \includegraphics[width=\linewidth]{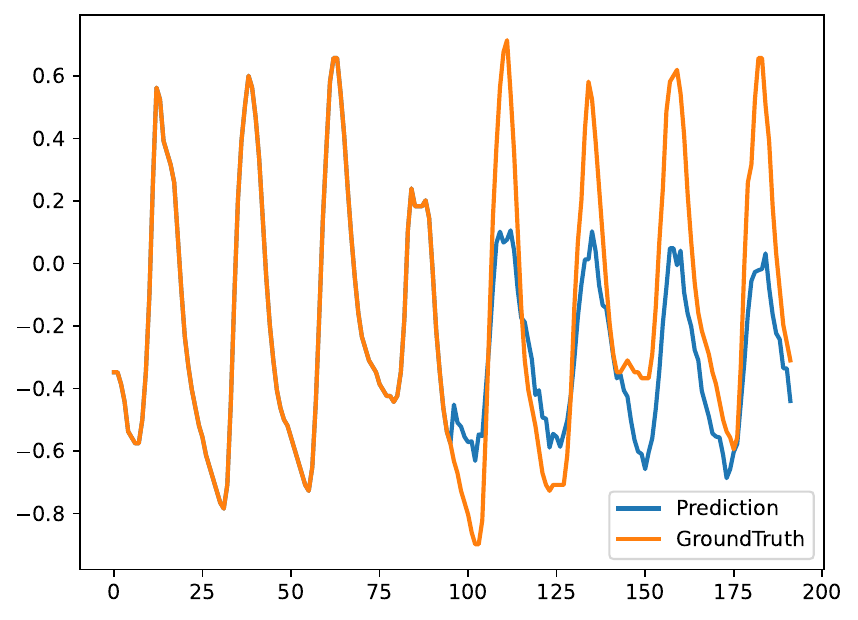}
            \centering
        \end{minipage} &
        \begin{minipage}{\imgwidth}
            \includegraphics[width=\linewidth]{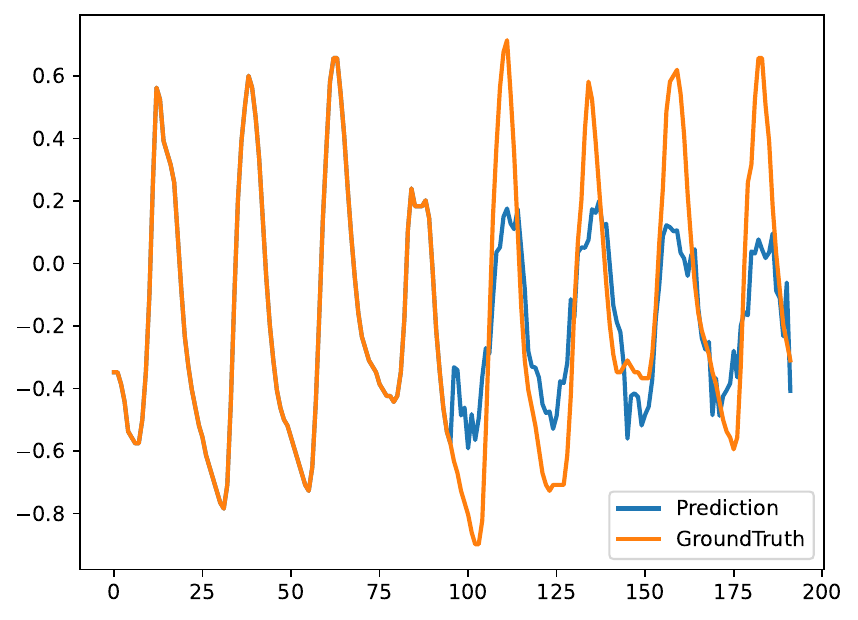}
            \centering
        \end{minipage} &
        \begin{minipage}{\imgwidth}
            \includegraphics[width=\linewidth]{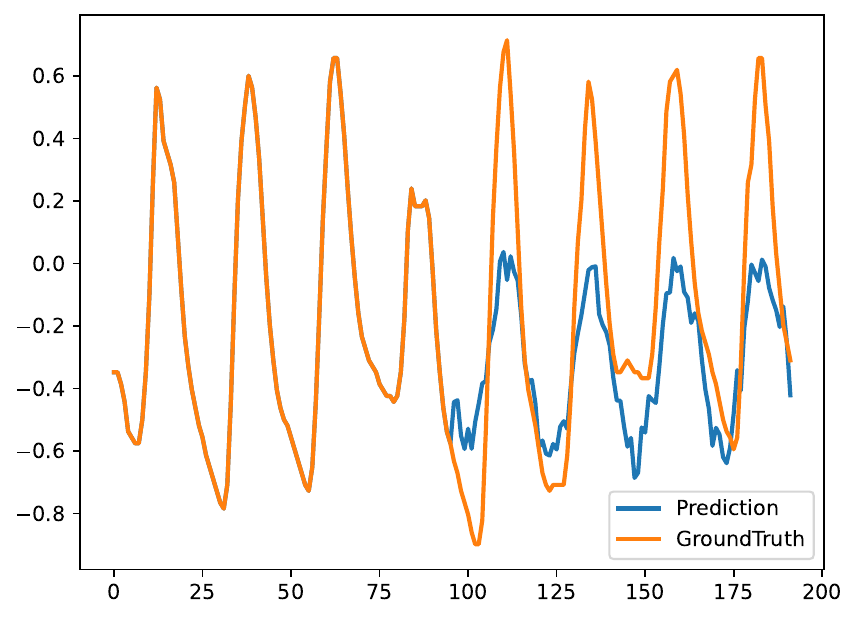}
            \centering
        \end{minipage} \\[0.5em]
        
        & \small (a) SCPaT & \small (b) PatchTST & \small (c) TimesNet & \small (d) iTransformer \\
    \end{tabular}
    \vspace{0.5em}
    \caption{Visualization of prediction results on Electricity, Traffic, and ETTh2 under the Lookback-96-Horizon-96 setting.}
    \label{fig6}
\end{figure*}

\subsection{Case Study and Visualization}

\subsubsection{Adjacency Matrix Interpretation.}

To further investigate the structural dependencies captured by SCPaT, we visualize the final adjacency matrix \(A_{\text{final}}\) produced by the semantic graph constructor under \(L=96\) and \(H=96\), as shown in Figure~\ref{fig5}. We present ETTh1 and Weather as representative cases, as their learned adjacency matrices exhibit markedly different structural patterns.
On ETTh1, the adjacency matrix exhibits a clear block structure, with strong interactions both within variable specific regions and across several groups of semantic units. This pattern indicates that SCPaT captures both local temporal dependencies and cross variable interactions. In contrast, the adjacency matrix on Weather is more concentrated around the diagonal, with relatively sparse activations in distant regions. This pattern suggests a more localized dependency structure, where semantic units primarily interact with temporally or structurally nearby units. The distinct adjacency patterns across the two datasets further illustrate that SCPaT learns dataset dependent dependency structures rather than imposing a fixed interaction pattern.

\begin{table*}
\caption{Ablation results of the SCPaT components on ETTm2, Weather, and Traffic over prediction horizons $H \in \{96, 192, 336, 720\}$, with the input length fixed at $L=96$. The best and second-best results are marked in \textcolor{red}{\textbf{bold red}} and \textcolor{blue}{\underline{blue underline}}, respectively.}
\centering
\setlength{\tabcolsep}{3pt}
\renewcommand{\arraystretch}{1.2}
\scriptsize
\begin{tabular}{cccccccccccccc}
\toprule
\multicolumn{2}{c}{Dataset} 
& \multicolumn{4}{c}{ETTm2} 
& \multicolumn{4}{c}{Weather} 
& \multicolumn{4}{c}{Traffic} \\
\cmidrule(lr){3-6} \cmidrule(lr){7-10} \cmidrule(lr){11-14}
\multicolumn{2}{c}{Prediction Length} 
& 96 & 192 & 336 & 720 
& 96 & 192 & 336 & 720 
& 96 & 192 & 336 & 720 \\
\midrule
\multirow{2}{*}{SCPaT} 
& MSE &\textcolor{red}{\textbf{0.173}} &\textcolor{red}{\textbf{ 0.238}} & \textcolor{red}{\textbf{0.297}} &\textcolor{red}{\textbf{ 0.394}} & \textcolor{red}{\textbf{0.163} }&\textcolor{red}{\textbf{ 0.209} }& \textcolor{red}{\textbf{0.266}}& \textcolor{red}{\textbf{0.345}} &\textcolor{red}{\textbf{ 0.393 }}&\textcolor{red}{\textbf{ 0.425}} &\textcolor{red}{\textbf{ 0.454}} &\textcolor{red}{\textbf{ 0.489}}\\
& MAE &\textcolor{red}{\textbf{0.258}} &\textcolor{red}{\textbf{ 0.300}} & \textcolor{red}{\textbf{0.339}} &\textcolor{red}{\textbf{ 0.394} }&\textcolor{red}{\textbf{ 0.208}} &\textcolor{red}{\textbf{ 0.250} }&\textcolor{red}{\textbf{ 0.292}} & \textcolor{red}{\textbf{0.349}} & \textcolor{red}{\textbf{0.247}} &\textcolor{red}{\textbf{ 0.259}} &\textcolor{red}{\textbf{ 0.270}} &\textcolor{red}{\textbf{ 0.291}}  \\
\midrule
\multirow{2}{*}{w/o SV Encoder} 
& MSE & 0.196 &0.259 & 0.317 & 0.412 & 0.204 & 0.248 & 0.302 & 0.373 & 0.414 &0.448  &0.478  &0.510  \\ 
 & MAE & 0.278 & 0.318 & 0.354 & 0.408 & 0.254 & 0.287 & 0.323 & 0.367 &0.274  &0.281  &0.294  &0.316 \\
\midrule
\multirow{2}{*}{w/o IAR} 
& MSE &\textcolor{blue}{\underline{0.178}}  &\textcolor{blue}{\underline{0.245}} &\textcolor{blue}{\underline{0.302}}  &\textcolor{blue}{\underline{0.400}}  &\textcolor{blue}{\underline{0.167}}  &0.216 &\textcolor{blue}{\underline{0.273}}  &\textcolor{blue}{\underline{0.353 }}&\textcolor{blue}{\underline{0.397}}  &\textcolor{blue}{\underline{0.431}}  &\textcolor{blue}{\underline{0.459}}  &\textcolor{blue}{\underline{0.495}}\\ 
 & MAE &\textcolor{blue}{\underline{0.264}}  &\textcolor{blue}{\underline{0.309}}  &0.347 &\textcolor{blue}{\underline{0.402}}  & \textcolor{blue}{\underline{0.214}} & \textcolor{blue}{\underline{0.257}} & \textcolor{blue}{\underline{0.299}} &\textcolor{blue}{\underline{ 0.354}} & 0.253 &0.267 &\textcolor{blue}{\underline{0.274}} & \textcolor{blue}{\underline{0.297 }}\\
\midrule
\multirow{2}{*}{w/o TE Graph} 
& MSE&0.180  &0.247  &0.309  &0.407  &0.171  &\textcolor{blue}{\underline{0.214}}  &0.277  &0.356  &0.401  &0.436  &0.466  &0.498  \\  
 & MAE &0.270  &0.311  &\textcolor{blue}{\underline{0.343}}  &0.413  &0.221  &0.262  & 0.301 & 0.355 & \textcolor{blue}{\underline{0.249}} & \textcolor{blue}{\underline{0.264}} & 0.277 & 0.301\\
\bottomrule
\end{tabular}
\label{Ablation}
\end{table*}

\subsubsection{Visualization of Forecasting Results.}

Figure~\ref{fig6} provides a qualitative comparison between SCPaT and several representative baselines. We select three datasets with progressively increasing dimensionality, covering low, medium, and high dimensional forecasting scenarios, to examine whether the prediction quality remains consistent as the complexity of multivariate interactions increases.
Across the three examples, SCPaT produces predictions that more closely follow the ground truth. It better preserves the periodic structure and amplitude variation of the target sequence while accurately tracking local trend changes. In highly oscillatory regions, SCPaT also maintains the phase and waveform shape more effectively, whereas the baseline methods exhibit larger deviations, particularly around turning points and local extrema. These qualitative results are consistent with the quantitative improvements reported above and illustrate the ability of SCPaT to model temporal patterns across datasets with different dimensional scales.

\subsection{Ablation Study}

To evaluate the contribution of each major component in SCPaT, we conduct ablation studies on ETTm2, Weather, and Traffic, representing low, medium, and high dimensional forecasting settings, respectively. We consider three variants: \textbf{W/O SV Encoder}, which removes the Semantic Vector Encoder; \textbf{W/O IAR}, which removes the Importance Aware Routing mechanism; and \textbf{W/O TE Graph}, which removes the Transfer Entropy Graph Constructor. Table~\ref{Ablation} summarizes the results.
The complete SCPaT consistently achieves the best performance across the evaluated datasets and forecasting horizons, while removing any individual component leads to performance degradation. In particular, removing the Semantic Vector Encoder or Importance Aware Routing generally results in larger performance drops, demonstrating their substantial contributions to the overall forecasting performance. The degradation observed without the TE Graph further confirms the benefit of modeling directed dependencies among semantic units. These results indicate that the three components provide complementary contributions to SCPaT.

\subsection{Model Efficiency}

\begin{figure}  
\centering
\includegraphics[width=1.1\columnwidth]{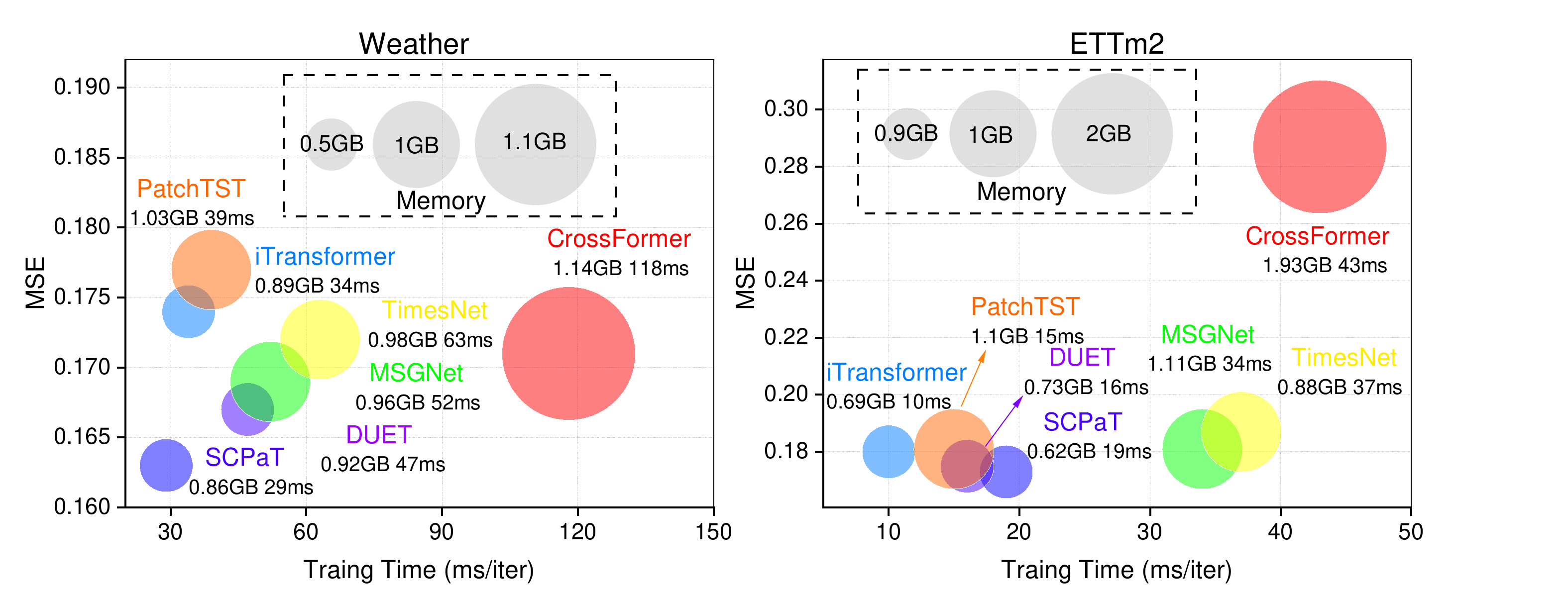}
\caption{Model efficiency comparison of different methods on ETTm2 and Weather.}
\label{fig7}
\end{figure}

We compare SCPaT with representative baselines in terms of forecasting accuracy, training time, and peak GPU memory usage under their recommended model configurations. Figure~\ref{fig7} illustrates the efficiency and performance trade-off on ETTm2 and Weather with the input length fixed at \(L=96\) and the prediction horizon fixed at \(H=96\). We select these two datasets as representative benchmarks with different dimensional scales and temporal characteristics, enabling a compact comparison under consistent experimental settings.
The horizontal axis denotes training time, the vertical axis represents MSE, and the bubble size indicates peak GPU memory consumption during training. SCPaT achieves competitive accuracy with moderate computational and memory costs, providing a favorable balance between performance and efficiency.

\section{Conclusion}
This paper revisits patch based MTSF from the perspective of semantic structure. Rather than relying solely on fixed, multiscale, or extendable partitioning schemes, we emphasize the importance of preserving semantic consistency within local temporal patterns while modeling higher order dependencies among them. Based on this perspective, we propose SCPaT, a unified framework that learns semantically coherent temporal units and their directed dependencies through a transfer entropy inspired formulation, and adaptively exploits the resulting structured representations for forecasting.
Extensive experiments across diverse forecasting settings demonstrate that SCPaT achieves strong and consistent performance compared with representative baselines. The results further indicate that incorporating semantic structure into temporal partitioning provides an effective way to model heterogeneous temporal patterns and complex multivariate dependencies. Future work may further explore the applicability of SCPaT to irregularly sampled time series forecasting.



\section*{Generative AI disclosure}

The authors used artificial intelligence tools (e.g., ChatGPT) solely for language polishing and grammar correction. All research concepts, methodology, and visual elements were developed independently without AI assistance. The authors take full responsibility for the scholarly content and academic integrity of the work.

\section*{Acknowledgments}
This work was supported in part by the Key Research Project Plan for Basic Research Special Fund of Higher Education Institutions in Henan Province under Grant No.25ZX009, and by the Henan Provincial Natural Science Foundation under Grant No.262300422534.

\printcredits

\bibliographystyle{cas-model2-names}

\bibliography{cas-refs}



\end{document}